%% file: 0896.tex
\documentclass[runningheads]{llncs}
\usepackage{graphicx}
\usepackage{comment}
\usepackage{amsmath,amssymb} 
\usepackage{color}
\usepackage{soul}
\usepackage{xcolor}
\usepackage{microtype}
\usepackage{wrapfig}
\usepackage{booktabs}       
\usepackage{multirow}
\usepackage{paralist}
\usepackage{subcaption}
\usepackage{pifont}

\usepackage[pdftex, pagebackref, colorlinks, citecolor=blue, linkcolor=blue]{hyperref}

\begin{document}
\pagestyle{headings}
\mainmatter
\def\ECCVSubNumber{896}  
\newcommand{\new}[1]{\textcolor{black}{#1}}

\title{Domain Adaptive Semantic Segmentation \\ Using Weak Labels} 

\titlerunning{Domain Adaptive Semantic Segmentation Using Weak Labels} 
\authorrunning{S. Paul, Y.-H. Tsai, S. Schulter, A. K. Roy-Chowdhury, M. Chandraker} 
\author{Sujoy Paul$^1$, Yi-Hsuan Tsai$^2$, Samuel Schulter$^2$, \\ Amit K. Roy-Chowdhury$^1$, Manmohan Chandraker$^{2,3}$}
\institute{$^1$UC Riverside $\ \ \ \ \ \ ^2$NEC Labs America $\ \ \ \ \ \ \ ^3$UC San Diego}

\maketitle

\begin{abstract}
    Learning semantic segmentation models requires a huge amount of pixel-wise labeling. However, labeled data may only be available abundantly in a domain different from the desired target domain, which only has minimal or no annotations. In this work, we propose a novel framework for domain adaptation in semantic segmentation with image-level weak labels in the target domain. The weak labels may be obtained based on a model prediction for unsupervised domain adaptation (UDA), or from a human annotator in a new weakly-supervised domain adaptation (WDA) paradigm for semantic segmentation. Using weak labels is both practical and useful, since (i) collecting image-level target annotations is comparably cheap in WDA and incurs no cost in UDA, and (ii) it opens the opportunity for category-wise domain alignment.
	Our framework uses weak labels to enable the interplay between feature alignment and pseudo-labeling, improving both in the process of domain adaptation. Specifically, we develop a weak-label classification module to enforce the network to attend to certain categories, and then use such training signals to guide the proposed category-wise alignment method. In experiments, we show considerable improvements with respect to the existing state-of-the-arts in UDA and present a new benchmark in the WDA setting. Project page is at \url{http://www.nec-labs.com/~mas/WeakSegDA}.
\end{abstract}

\input{intro.tex}
\input{related.tex}
\input{method.tex}

\input{experiment.tex}

\section{Conclusions}
In this paper, we use weak labels to improve domain adaptation for semantic segmentation in both the UDA and WDA settings, with the latter being a novel setting.
Specifically, we design an image-level classification module using weak labels, enforcing the network to pay attention to categories that are present in the image.
With such a guidance from weak labels, we further utilize a category-wise alignment method to improve adversarial alignment in the feature space.
Based on these two mechanisms, our formulation generalizes both to pseudo-weak and oracle-weak labels.
We conduct extensive ablation studies to validate our approach against state-of-the-art UDA approaches.

\noindent
\textbf{Acknowledgment.} This work was a part of Sujoy Paul's internship at NEC Labs America. This work was also partially funded by NSF grant 1724341. 

\clearpage
%
%
\bibliographystyle{splncs04}
\bibliography{egbib}

\input{supp}

\end{document}

%% file: intro.tex
\section{Introduction}

Unsupervised domain adaptation (UDA) methods for semantic segmentation have been developed to tackle the issue of domain gap.
Existing methods aim to adapt a model learned on the source domain with pixel-wise ground truth annotations, e.g., from a simulator which requires the least annotation efforts, to the target domain that does not have any form of annotations.
These UDA methods in the literature for semantic segmentation are developed mainly using two mechanisms: pseudo label self-training and distribution alignment between the source and target domains.
For the first mechanism, pixel-wise pseudo labels are generated via strategies such as confidence scores \cite{Li_CVPR_2019,Hung_BMVC_2018} or self-paced learning \cite{Zou_ECCV_2018}, but such pseudo-labels are specific to the target domain, and do not consider alignment between domains.
For the second mechanism, numerous spaces could be considered to operate the alignment procedure, such as pixel \cite{Hoffman_ICML_2018,Murez_CVPR_2018}, feature \cite{Hoffman_CoRR_2016,Zhang_ICCV_2017}, output \cite{tsai2018learning,Chen_CVPR_2018}, and patch \cite{Tsai_DA4Seg_ICCV19} spaces.
However, alignment performed by these methods are agnostic to the category, which may be problematic as the domain gap may vary across categories.
\begin{figure}[t]
    \centering
    \includegraphics[width=1.0\textwidth]{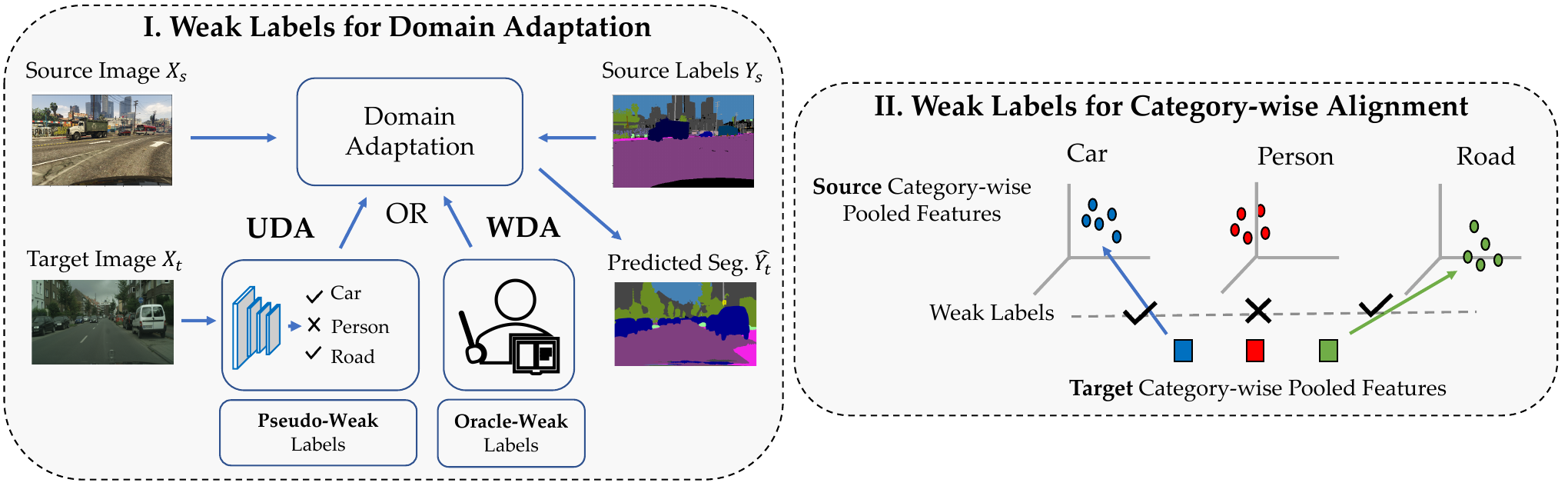}
    \caption{Our work introduces two key ideas to adapt semantic segmentation models across domains. I: Using image-level weak annotations for domain adaptation, either estimated, i.e., pseudo-weak labels (Unsupervised Domain Adaptation, UDA) or acquired from a human oracle (Weakly-supervised Domain Adaptation (WDA). II: We utilize weak labels to improve the category-wise feature alignment between the source and target domains. \checkmark/\ding{53} depicts weak labels, i.e., the categories present/absent in an image.}
    \label{fig:teaser}
\end{figure}

To alleviate the issue of lacking annotations in the target domain, we propose a concept of utilizing \emph{weak labels} on the domain adaptation task for semantic segmentation, in the form of image- or point-level annotations in the target domain. Such weak labels can be used for category-wise alignment between the source and target domain, and also to enforce constraints on the categories present in an image.
It is important to note that our weak labels could be estimated from the model prediction in the UDA setting, or provided by the human oracle in the weakly-supervised domain adaptation (WDA) paradigm (see left of Fig. \ref{fig:teaser}).
We are the first to introduce the WDA setting for semantic segmentation with image-level weak-labels, which is practically useful as collecting such annotations is much easier than pixel-wise annotations on the target domain.
Benefiting from the concept of weak labels introduced in this paper, we aim to utilize such weak labels to act as an enabler for the interplay between the alignment and pseudo labeling procedures, as they are much less noisy compared to pixel-wise pseudo labels.
Specifically, we use weak labels to perform both 1) image-level classification to identify the presence/absence of categories in an image as a regularization, and 2) category-wise domain alignment using such categorical labels. 
For the image-level classification task, weak labels help our model obtain a better pixel-wise attention map per category.
Then, we utilize the category-wise attention maps as the guidance to further pool category-wise features for proposed domain alignment procedure (right of Fig. \ref{fig:teaser}).
\new{
Note that, although weak labels have been used in domain adaptation for object detection \cite{inoue2018cross}, our motivation is different from theirs. More specifically, \cite{inoue2018cross} uses the weak labels to choose pseudo labels for self-training, while we formulate a general framework to learn from weak labels with different forms, i.e., UDA and WDA (image-level or point supervision), as well as to improve feature alignment across domains using weak labels.
}
We conduct experiments on the road scene segmentation problem from GTA5~\cite{Richter_ECCV_2016}/SYNTHIA~\cite{Ros_CVPR_2016} to Cityscapes~\cite{cityscapes}.
We perform extensive experiments to verify the usefulness of each component in the proposed framework, and show that our approach performs favorably against state-of-the-art algorithms for UDA.
In addition, we show that our proposed method can be used for WDA and present its experimental results as a new benchmark.
\new{For the WDA setting, we also show that our method can incorporate various types of weak labels, such as image-level or point supervision.}
The \textbf{main contributions} of our work are: 1) we propose a concept of using weak labels to help domain adaptation for semantic segmentation; 2) we utilize weak labels to improve category-wise alignment for better feature space adaptation; and 3) we demonstrate that our method is applicable to both UDA and WDA settings.

%% file: related.tex
\section{Related Work}

In this section, we discuss the literature of unsupervised domain adaptation (UDA) for image classification and semantic segmentation. In addition, we also discuss weakly-supervised methods for semantic segmentation.

{\flushleft \bf{UDA for Image Classification.}}
The UDA task for image classification has been developed via aligning distributions across source and target domains.
To this end, hand-crafted features \cite{Fernando_ICCV_2013,gong2012geodesic} and deep features \cite{ganin2015unsupervised,tzeng2015simultaneous} have been considered to minimize the domain discrepancy and learn domain-invariant features.
To further enhance the alignment procedure, maximum mean discrepancy~\cite{long2015learning} and adversarial learning~\cite{ganin2016domain,tzeng2017adversarial} based approaches have been proposed.
Recently, several algorithms focus on improving deep models \cite{long2016unsupervised,Saito_CVPR_2018,Lee_CVPR_2019,dai2019adaptation}, combining distance metric learning \cite{sohn2017unsupervised,sohn2019unsupervised}, utilizing pixel-level adaptation \cite{Bousmalis_CVPR_2017,Luan_CVPR_2019}, or incorporating active learning \cite{Su_WACV_2020}.
{\flushleft \bf{UDA for Semantic Segmentation.}}
Existing UDA methods in literature for semantic segmentation can be categorized primarily into to two groups: domain alignment and pseudo-label self-training.
For domain alignment, numerous algorithms focus on aligning distributions in the pixel \cite{Chang_CVPR_2019,Choi_ICCV19,Hoffman_ICML_2018,Murez_CVPR_2018,Wu_ECCV_2018,Zhang_CVPR_2018}, feature \cite{Chen_ICCV_2017,Hoffman_CoRR_2016,Zhang_ICCV_2017}, and output \cite{tsai2018learning,Chen_CVPR_2018} spaces.
For pseudo-label re-training, current methods~\cite{Saleh_ECCV_2018,Zou_ECCV_2018,Lian_ICCV19} aim to generate pixel-wise pseudo labels on the target images, which is utilized to finetune the segmentation model trained on the source domain.

To achieve better performance, recent works \cite{Du_ICCV19,Li_CVPR_2019,Tsai_DA4Seg_ICCV19,Vu_CVPR_2019} attempt to combine the above two mechanisms.
AdvEnt \cite{Vu_CVPR_2019} adopts adversarial alignment and self-training in the entropy space, while BDL \cite{Li_CVPR_2019} combines output space and pixel-level adaptation with pseudo-label self-training in an iterative updating scheme.
Moreover, Tsai et al. \cite{Tsai_DA4Seg_ICCV19} propose a patch-level alignment method and show that their approach is complementary to existing modules such as output space adaptation and pseudo-label self-training.
Similarly, Du et al. \cite{Du_ICCV19} integrate category-wise adversarial alignment with pixel-wise pseudo-labels, which may be noisy, leading to incorrect alignment.
In addition, \cite{Du_ICCV19} needs to progressively change a ratio for selecting pseudo-labels, and the final performance is sensitive to this chosen parameter.

Compared to the above-mentioned approaches, we propose to exploit weak labels by learning an image classification task, while improving domain alignment through category-wise attention maps.
Furthermore, we show that our approach can be utilized even in the case where oracle-weak labels are available on the target domain, in which case the performance will be further improved. 
{\flushleft \bf{Weakly-supervised Semantic Segmentation.}}
In this paper, since we are specifically interested in how weak labels can help domain adaptation, we also discuss the literature for weakly-supervised semantic segmentation, which has been tackled through different types of weak labels, such as image-level~\cite{ahn2018learning,Chang_CVPR_2020_WSSS,Kolesnikov_ECCV16,pathak2015constrained,Pinheiro_CVPR15}, video-level \cite{Chen_IJCV_2020,Tsai_ECCV_2016,Zhong_ACCV_2016}, bounding box \cite{papandreou2015weakly,dai2015boxsup,khoreva_CVPR17}, scribble \cite{Lin_CVPR16,Vernaza_CVPR17}, and point \cite{Bearman_ECCV16} supervisions.
Under this setting, these methods train the model using ground truth weak labels and perform testing in the same domain, which does not require domain adaptation.
In contrast, we use a source domain with pixel-wise ground truth labels, but in the target domain, we consider pseudo-weak labels (UDA) or oracle-weak labels (WDA).
As a result, we note that performance of weakly-supervised semantic segmentation methods which do not utilize any source domain, is usually much lower than the domain adaptation setting adopted in this paper, e.g., the mean IoU on Cityscapes is only 24.9$\%$ as shown in \cite{Saleh_ICCV17}.

%% file: method.tex
\section{Domain Adaptation with Weak Labels} \label{sec:method}

\begin{figure}[t]
    \centering
    
    \includegraphics[width=1.0\textwidth]{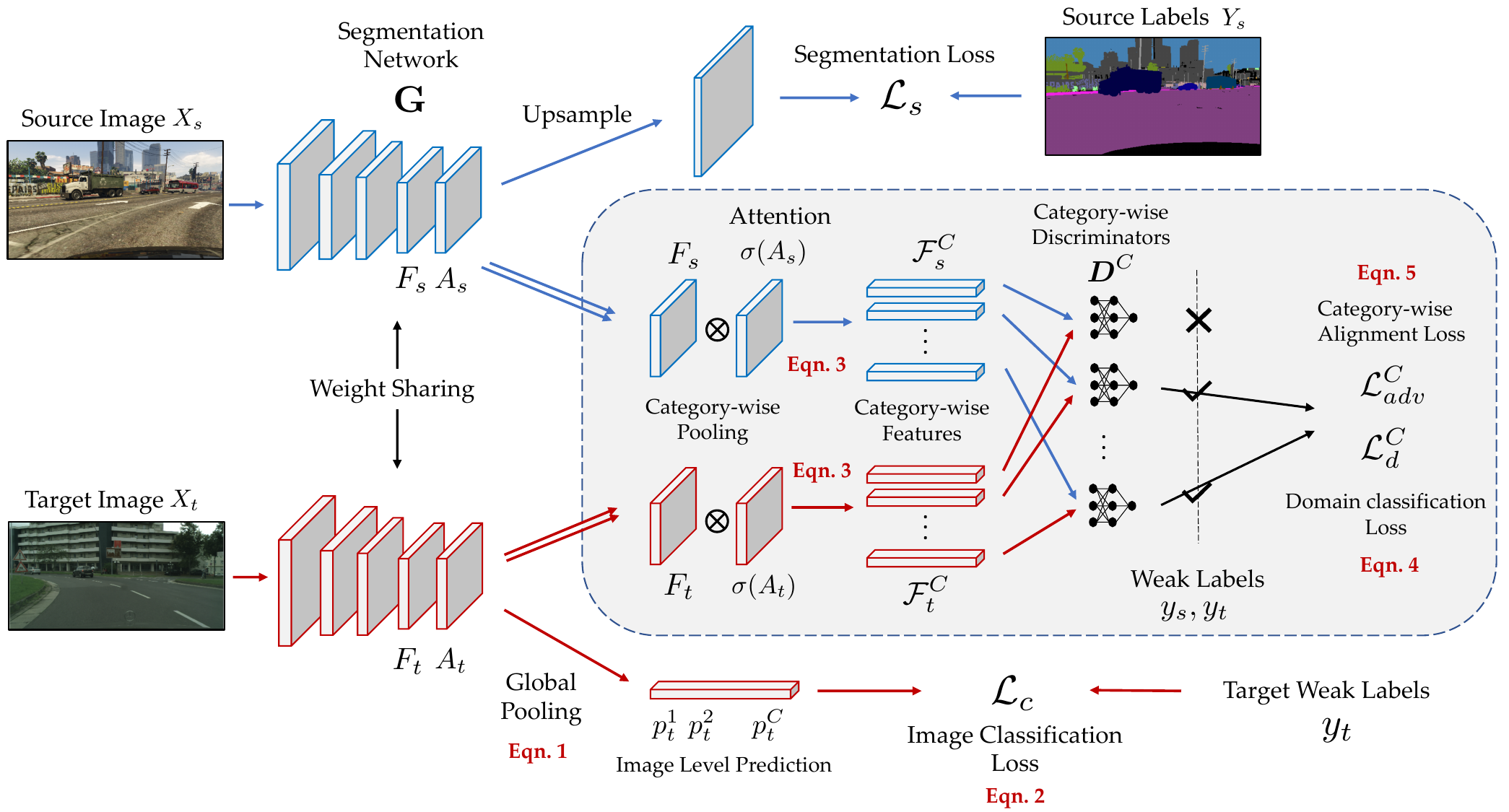}
    \caption{The proposed architecture consists of the segmentation network $\boldsymbol{G}$ and the weak label module. We compute the pixel-wise segmentation loss $\mathcal{L}_s$ for the source images and image classification loss $\mathcal{L}_c$ using the weak labels $y_t$ for the target images. Note that the weak labels can be estimated as pseudo-weak labels or provided by a human oracle. We then use the output prediction $A$, convert it to an attention map $\sigma(A)$ and pool category-wise features $\mathcal{F}^C$. Next, these features are aligned between source and target domains using the category-wise alignment loss $\mathcal{L}^C_{adv}$ guided by the category-wise discriminators $\boldsymbol{D}^C$ learned via the domain classification loss $\mathcal{L}_d^C$.}
    \label{fig:framework}
\end{figure}

In this section, we first precisely describe the problem statement we aim to solve in this work. We then provide a brief overview of our method and then explain the details of our framework - the image-level classification module and category-wise alignment method using weak labels. Finally, we present our method of obtaining the weak labels for the UDA and WDA settings. Note that at the end we also introduce using a different form of weak labels, the point labels, which further boosts the performance of our framework with little annotation cost.

\subsection{Problem Definition} \label{sec:problem}
In the source domain, we have images and pixel-wise labels denoted as $\mathcal{I}_s=\{{X}_s^i, {Y}_s^i\}_{i=1}^{N_s}$.  Whereas, our target dataset contains images and only image-level labels as $\mathcal{I}_t=\{{X}_t^i, {y}_t^i\}_{i=1}^{N_t}$. Note that ${X}_s, {X}_t \in \mathbb{R}^{H\times W\times 3}$, ${Y}_s \in \mathbb{B}^{H\times W\times C}$ with pixel-wise one-hot vectors, ${y}_t \in \mathbb{B}^C$ is a multi-hot vector representing the categories present in the image and $C$ is the number of categories, same for both the source and target datasets.
Such image-level labels ${y}_t$ are often termed as weak labels. We can either estimate them, in which case we call them pseudo-weak labels (Unsupervised Domain Adptation, UDA) or acquire them from a human oracle that is called oracle-weak labels (Weakly-supervised Domain Adaptation, WDA). 
We will further discuss details of acquiring weak labels in Section \ref{sec:uda_weak}. Given such data, the problem is to adapt a segmentation model $\boldsymbol{G}$ learned on the source dataset $\mathcal{I}_s$ to the target dataset $\mathcal{I}_t$.

\subsection{Algorithm Overview} \label{sec:overview}
Fig. \ref{fig:framework} presents an overview of our proposed method. We first pass both the source and target images through the segmentation network ${G}$ and obtain their features ${F}_s, {F}_t \in \mathbb{R}^{H'\times W'\times 2048}$, segmentation predictions ${A}_s, {A}_t \in \mathbb{R}^{H'\times W'\times C}$, and the up-sampled pixel-wise predictions ${O}_s, {O}_t \in \mathbb{R}^{H\times W \times C}$. Note that $H'(<H), W'(<W)$ are the downsampled spatial dimensions of the image after passing through the segmentation network. 
As a baseline, we use the source pixel-wise annotations to learn $\boldsymbol{G}$, while aligning the output space distribution ${O}_s$ and ${O}_t$, following \cite{tsai2018learning}.

In addition to having pixel-wise labels on the source data, we also have image-level weak labels on the target data. As discussed before, such weak labels can be either estimated (UDA) or acquired from an oracle (WDA).
We then utilize these weak labels to update the segmentation network $\boldsymbol{G}$ in two different ways. First, we introduce a module which learns to predict the categories that are present in a target image. Second, we formulate a mechanism to align the features of each individual category between source and target domains.
To this end, we use category-specific domain discriminators  $\boldsymbol{D}^{\boldsymbol{c}}$ guided by the weak labels to determine which categories should be aligned.
In the following sections, we present these two modules in more detail.

\subsection{Weak Labels for Category Classification} \label{sec:weak_classify}

\new{In order to predict whether a category is absent/present in a particular image, we define an image classification task using the weak labels, such that the segmentation network $\boldsymbol{G}$ can discover those categories.}
Specifically, we use the weak labels $y_t$ and learn to predict the categories present/absent in the target images. We first feed the target images $X_t$ through $\boldsymbol{G}$ to obtain the predictions $A_t$ and then apply a global pooling layer to obtain a single vector of predictions for each category:
\begin{align}
    p_t^c = \sigma_s \left[\frac{1}{k}\log\frac{1}{H'W'}\sum_{h',w'} \exp kA_t^{(h',w',c)} \right],
    \label{eq:pool}
\end{align}
where $\sigma_s$ is the sigmoid function such that $p_t$ represents the probability that a particular category appears in an image. Note that \eqref{eq:pool} is a smooth approximation of the \texttt{max} function. The higher the value of $k$, the better it approximates to \texttt{max}. We set $k=1$ as we do not want the network to focus only on the maximum value of the prediction, which may be noisy, but also on other predictions that may have high values.
Using $p_t$ and the weak labels $y_t$, we can compute the category-wise binary cross-entropy loss:
\begin{equation}
    \mathcal{L}_c(X_t; \boldsymbol{G}) = \sum_{c=1}^C - y_t^c\log(p_t^c) - (1-y_t^c)\log(1-p_t^c).
    \label{eq:bce}
\end{equation}
This is shown at the bottom stream of Fig. \ref{fig:framework}. This loss function $\mathcal{L}_c$ helps to identify the categories which are absent/present in a particular image and enforces the segmentation network $\boldsymbol{G}$ to pay attention to those objects/stuff that are partially identified when the source model is used directly on the target images.

\subsection{Weak Labels for Feature Alignment} \label{sec:weak_align}
The classification loss using weak labels introduced in \eqref{eq:bce} regularizes the network focusing on certain categories.
However, distribution alignment across the source and target domains is not considered yet. As discussed in the previous section, methods in literature either align feature space \cite{Hoffman_CoRR_2016} or output space \cite{tsai2018learning} across domains. However, such alignment is agnostic to the category, so it may align features of categories that are not present in certain images. Moreover, features belonging to different categories may have different domain gaps. Thereby, performing category-wise alignment could be beneficial but has not been widely studied in UDA for semantic segmentation.
Although an existing work \cite{Du_ICCV19} attempts to align category-wise features, it utilizes pixel-wise pseudo labels, which may be noisy, and performs alignment in a high-dimensional feature space, which is not only difficult to optimize but also requires more computations (more discussions are provided in the experimental section).

To alleviate all the above issues, we use image-level weak labels to perform category-wise alignment in the feature space. Specifically, we obtain the category-wise features for each image via an attention map, i.e., segmentation prediction, guided by our classification module using weak labels, and then align these features between the source and target domains.
We next discuss the category-wise feature pooling mechanism followed by the adversarial alignment technique. 

{\flushleft \bf{Category-wise Feature Pooling.}}
Given the last layer features $F$ and the segmentation prediction $A$, we obtain the category-wise features by using the prediction as an attention over the features. Specifically, we obtain the category-wise feature $\mathcal{F}^c$ as a 2048-dimensional vector for the $c^{th}$ category as follows:
\begin{equation}
    \mathcal{F}^c= \sum_{h',w'} \sigma (A)^{(h',w',c)} F^{(h',w')},
    \label{eq:pool_fea}
\end{equation}
\noindent
where $\sigma(A)$ is a tensor of dimension $H' \times W' \times C$, with each channel along the category dimension representing the category-wise attention obtained by the softmax operation $\sigma$ over the spatial dimensions. As a result, $\sigma(A)^{(h',w',c)}$ is a scalar and $F^{(h',w')}$ is a 2048-dimensional vector, while $\mathcal{F}^c$ is the summed feature of $F^{(h',w')}$ weighted by $\sigma(A)^{(h',w',c)}$ over the spatial map $H' \times W'$.
Note that we drop the subscripts $s,t$ for source and target, as we employ the same operation to obtain the category-wise features for both domains. We next present the mechanism to align these features across domains. Note that we will use $\mathcal{F}^c$ to denote the pooled feature for the $c^{th}$ category and $\mathcal{F}^C$ to denote the set of pooled features for all the categories. Category-wise feature pooling is shown in the middle of Fig. \ref{fig:framework}. 

{\flushleft \bf{Category-wise Feature Alignment.}}
To learn the segmentation network $\boldsymbol{G}$ such that the source and target category-wise features are aligned, we use an adversarial loss while using category-specific discriminators $\boldsymbol{D}^C=\{\boldsymbol{D}^c\}_{c=1}^C$.
\new{The reason of using category-specific discriminators is to ensure that the feature distribution for each category could be aligned independently, which avoids the noisy distribution modeling from a mixture of categories.}
In practice, we train $C$ distinct category-specific discriminators to distinguish between category-wise features drawn from the source and target images. The loss function to train the discriminators $\boldsymbol{D}^C$ is as follows:
\begin{equation}
    \mathcal{L}^C_d(\mathcal{F}_s^C,\mathcal{F}_t^C;\boldsymbol{D}^C)= \sum_{c=1}^C -y^c_s\log \boldsymbol{D}^c\big(\mathcal{F}_s^c\big)
    - y^c_t\log \big(1- \boldsymbol{D}^c\big(\mathcal{F}_t^c\big)\big).
    \label{eq:disc_loss}
\end{equation}
Note that, while training the discriminators, we only compute the loss for those categories which are present in the particular image via the weak labels $y_s, y_t \in \mathbb{B}^C$ that indicate whether a category occurs in an image or not.
Then, the adversarial loss for the target images to train the segmentation network $\boldsymbol{G}$ can be expressed as follows:
\begin{equation}
    \mathcal{L}^C_{adv}(\mathcal{F}_t^C;\boldsymbol{G},\boldsymbol{D}^C)= \sum_{c=1}^C -y_t^c\log \boldsymbol{D}^c\big(\mathcal{F}_t^c\big).
    \label{eq:adv_loss}
\end{equation}
Similarly, we use the target weak labels $y_t$ to align only those categories present in the target image.
By minimizing $\mathcal{L}_{adv}^C$, the segmentation network tries to fool the discriminator by maximizing the probability of the target category-wise feature being considered as drawn from the source distribution. These loss functions in \eqref{eq:disc_loss} and \eqref{eq:adv_loss} are obtained in the right of the middle box in Fig. \ref{fig:framework}.

\subsection{Network Optimization} \label{sec:optim}
{\flushleft \bf{Discriminator Training.}}
We learn a set of $C$ distinct discriminators for each category $c$. We use the source and target images to train the discriminators, which learn to distinguish between the category-wise features drawn from either the source or the target domain. The optimization problem to train the discriminator can be expressed as: $\min_{\boldsymbol{D}^C} \mathcal{L}_d^C(\mathcal{F}_s^C, \mathcal{F}_t^C).$
Note that each discriminator is trained only with features pooled specific to that particular category. Therefore, given an image, we only update those discriminators corresponding to those categories which are present in the image and ignore the rest.

{\flushleft \bf{Segmentation Network Training.}}
We train the segmentation network with the pixel-wise cross-entropy loss $\mathcal{L}_s$ on the source images, image classification loss $\mathcal{L}_c$ and adversarial loss $\mathcal{L}_{adv}^C$ on the target images.
We combine these loss functions to learn $\mathbf{G}$ as follows :
\begin{equation}
    \min_{\boldsymbol{G}} \mathcal{L}_s(X_s) + \lambda_c\mathcal{L}_c(X_t) + \lambda_d\mathcal{L}_{adv}^C(\mathcal{F}_t^C).
    \label{eq:joint_train}
\end{equation}
We follow the standard GAN training procedure \cite{Goodfellow_NIPS_2014} to alternatively update $\boldsymbol{G}$ and $\boldsymbol{D}^C$.
Note that, computing $\mathcal{L}_{adv}^C$ involves the category-wise discriminators $\boldsymbol{D}^C$. Therefore, we fix $\boldsymbol{D}^C$ and backpropagate gradients only for the segmentation network $\boldsymbol{G}$.

\subsection{Acquiring Weak Labels} \label{sec:uda_weak}
In the above sections, we have proposed a mechanism to utilize image-level weak labels of the target images and adapt the segmentation model between source and target domains. In this section, we explain two methods to obtain such image-level weak labels.

{\flushleft \bf{Pseudo-Weak Labels (UDA).}} One way of obtaining weak labels is to directly estimate them using the data we have, i.e., source images/labels and target images, which is the unsupervised domain adaptation (UDA) setting.
In this work, we utilize the baseline model \cite{tsai2018learning} to adapt a model learned from the source to the target domain, and then obtain the weak labels of the target images as follows:
\begin{equation}
    y_t^c = \begin{cases}
      1, & \text{if}\ p^c_t > T, \\
      0, & \text{otherwise}
    \end{cases}
    \label{eq:pseudo_weak}
\end{equation}
where $p^c_t$ is the probability for category $c$ as computed in \eqref{eq:pool} and $T$ is a threshold, which we set to $0.2$ in all the experiments unless specified otherwise.
In practice, we compute the weak labels online during training and avoid any additional inference step. Specifically, we forward a target image, obtain the weak labels using \eqref{eq:pseudo_weak}, and then compute the loss functions in \eqref{eq:joint_train}.
As the weak labels obtained in this manner do not require human supervision, adaptation using such labels is unsupervised.

{\flushleft \bf{Oracle-Weak Labels (WDA).}} In this form, we obtain the weak labels by querying a human oracle to provide a list of the categories that occur in the target image. As we use supervision from an oracle on the target images, we refer to this as weakly-supervised domain adaptation (WDA). 
It is worth mentioning that the WDA setting could be practically useful, as collecting such human annotated weak labels is much easier than pixel-wise annotations. Also, there has not been any prior research involving this setting for domain adaptation. 

To show that our method can use different forms of oracle-weak labels, we further introduce the point supervision as in \cite{Bearman_ECCV16}, which only increases effort by a small amount compared to the image-level supervision. In this scenario, we randomly obtain one pixel coordinate of each category that belongs in the image, i.e., the set of tuples $ \{(h^c, w^c, c)| \forall y^c_t=1 \} $. For an image, we compute the loss as follows: $\mathcal{L}_{point} = - \sum_{\forall y^c_t=1} y^c_t \log(O_t^{(h^c, w^c, c)})$, where $O_t \in \mathbb{R}^{H\times W\times C}$ is the output prediction of target after pixel-wise softmax.

%% file: experiment.tex
\section{Experimental Results}

In this section, we perform an evaluation of our domain adaptation framework for semantic segmentation. We present the results for using both pseudo-weak labels, i.e., unsupervised domain adaptation (UDA) and human oracle-weak labels, i.e., weakly-supervised domain adaptation (WDA) and compare it with existing state-of-the-art methods. We also perform ablation studies to analyse the benefit of using pseudo/oracle-weak labels via our proposed weak-label classification module and category-wise alignment. 

{\flushleft \bf{Datasets and Metric.}}
We evaluate our domain adaptation method under the Sim-to-Real case with two different source-target scenarios. First, we adapt from GTA5 \cite{Richter_ECCV_2016} to the Cityscapes dataset \cite{cityscapes}. Second, we use SYNTHIA \cite{Ros_CVPR_2016} as the source and Cityscapes as the target, which has a larger domain gap than the former case. For all experiments, we use the Intersection-over-Union (IoU) ratio as the metric. For SYNTHIA$\rightarrow$Cityscapes, following the literature \cite{Vu_CVPR_2019}, we report the performance averaged over $16$ categories (listed in Table \ref{table:synthia_all}) and $13$ categories (removing wall, fence and pole), which we denote as mIoU*.

{\flushleft \bf{Network Architectures.}}
For the segmentation network $\boldsymbol{G}$, to have a fair comparison with works in literature, we use the DeepLab-v2 framework \cite{deeplab} with the ResNet-101 \cite{He_CVPR_2016} architecture. We extract features $F_s,F_t$ before the Atrous Spatial Pyramid Pooling (ASPP) layer. For the category-wise discriminators $D^C=\{D^c\}_{c=1}^C$, we use $C$ separate networks, where each consists of three fully-connected layers, having number of nodes $\{2048, 2048, 1\}$ with ReLU activation.

{\flushleft \bf{Training Details.}} We implement our framework using PyTorch on a single Titan X GPU with 12G memory for all our experiments. We use the SGD method to optimize the segmentation network and the Adam optimizer \cite{Kingma_ICLR_2015} to train the discriminators. We set the initial learning rates to be $2.5\times 10^{-4}$ and $1\times 10^{-4}$ for the segmentation network and discriminators, with polynomial decay of power 0.9 \cite{deeplab}.
As a common practice in weakly-supervised semantic segmentation \cite{ahn2018learning}, we use Dropout of $0.1$ and $0.3$ for oracle-weak labels and pseudo-weak labels respectively, on the spatial predictions before computing the loss $\mathcal{L}_c$.
We choose $\lambda_c$ to be $0.2$ for oracle-weak labels and use a smaller $\lambda_c = 0.01$ for pseudo-weak labels to account for its inaccurate prediction.
For the weight on the category-wise adversarial loss $\mathcal{L}_{adv}^C$, we set $\lambda_{adv}=0.001$.
For experiments using pseudo weak labels, to avoid noisy pseudo weak label prediction in the early training stage, we first train the segmentation baseline network using \cite{tsai2018learning} for 60K iterations.
Then, we include the proposed weak-label classification and alignment procedure, and train the entire framework. 

\subsection{Comparison with State-of-the-art Methods} \label{sec:sota}

{\flushleft \bf{Unsupervised Domain Adaptation (UDA).}}
We compare our method with existing state-of-the-art UDA methods in Table \ref{table:gta5_all} for GTA5$\rightarrow$Cityscapes and in Table \ref{table:synthia_all} for SYNTHIA$\rightarrow$Cityscapes.
Recent methods \cite{Chang_CVPR_2019,Hoffman_ICML_2018,Li_CVPR_2019,Tsai_DA4Seg_ICCV19} show that adapting images from source to target on the pixel level and then adding those translated source images in training enhances the performance.
We follow this practice in the final model via adding these adapted images to the source dataset, as their pixel-wise annotations do not change after adaptation.
Thus adaptation using weak labels aligns the features not only between the original source and target images, but also between the translated source images and the target images.
We show that our method is also complementary to pixel-level adaptation.

{\flushleft \bf{Discussions.}}
\new{
In terms of applied techniques, e.g, pseudo-label re-training and domain alignment, the closest comparisons to our method are DISE~\cite{Chang_CVPR_2019}, BDL~\cite{Li_CVPR_2019}, and Patch Space alignment~\cite{Tsai_DA4Seg_ICCV19}. We show that our method performs favorably against these approaches on both benchmarks.
This can be attributed to our introduced concept of using weak labels, in which our UDA model explores pseudo-weak image-level labels, instead of using pixel-level pseudo-labels \cite{Tsai_DA4Seg_ICCV19,Li_CVPR_2019} that may be noisy and degrade the performance.
In addition, these methods do not perform domain alignment guided by such pseudo labels, whereas we use weak labels to enable our category-wise alignment procedure.}

\new{
The only prior work that adopts category-wise feature alignment is SSF-DAN~\cite{Du_ICCV19}. However, our method is different from theirs in three aspects: 1) We introduce the weak-label classification module to take advantage of image-level weak labels that enables an efficient feature alignment process and the novel WDA setting;
2) Our unified framework can be applied for both UDA and WDA settings with various types of supervisions;
3) Due to the introduced weak-label module, our category-wise feature alignment is operated in the pooled feature space in \eqref{eq:pool_fea} guided by an attention map, rather than in a much higher-dimensional spatial space as in \cite{Du_ICCV19} that uses pixel-wise pseudo-labels. This essentially improves the training efficiency compared to \cite{Du_ICCV19}, which requires a GPU with 16 GB memory as their discriminator needs much more computation time ($>20 \times$) and GPU memory ($>8 \times$) compared to our combined output space and category-wise discriminators. Also, the discriminators in \cite{Du_ICCV19} require 130 GFLOPS, whereas our discriminators require a total of only 0.5 GFLOPS.
}

\input{tables/table_gta5_cityscapes.tex}

\subsection{Weakly-supervised Domain Adaptation (WDA)} \label{sec:sota}

{\flushleft \bf{Image-level Supervision.}}
We present the results of our method when using oracle-weak labels (obtained from the ground truth of the training set) in the last rows of Table \ref{table:gta5_all} and \ref{table:synthia_all}. To the best of our knowledge, we are the first to work on WDA, i.e., using human oracle-weak labels on domain adaptation for semantic segmentation, and there are no other methods to compare against in the literature. From the results, it is interesting to note that the major boost in performance using WDA compared to UDA occurs for categories such as truck, bus, train, and motorbike for both cases using GTA5 and SYNTHIA as the source domain. One reason is that those categories are most underrepresented in both the source and the target datasets. Thus, they are not predicted in most of the target images, but using the oracle-weak labels helps to identify them better.

{\flushleft \bf{Point Supervision.}}
We introduce another interesting setting of point supervision as in \cite{Bearman_ECCV16}, which adds only a slight increase of annotation time compared to the image-level supervision.
We follow \cite{Bearman_ECCV16} and randomly sample one pixel per category in each target image as the supervision.
Note that, all the details and the modules are the same during training in this setting.
In Table \ref{table:gta5_all} and \ref{table:synthia_all}, the results show that using point supervision improves performance (\textbf{$3.4-6.6\%$}) on both benchmarks compared to the image-level supervision.
This shows that our method is a general framework that can be applied to the conventional UDA setting as well as the WDA setting using either image-level or point supervision, while all the settings achieve consistent performance gains. 

Fig. \ref{fig:perf_comp} shows a comparison of annotation time v.s. performance for various levels of supervision. With low annotation cost in WDA cases, our model bridges the gap in performance between UDA and full supervision ones (more results are shown in the supplementary material). Note that, other forms of weak labels such as object count and density can also be effective. 

\input{tables/table_synthia_cityscapes.tex}

\input{tables/table_ablation.tex}

\subsection{Ablation Study} \label{sec:ablation}

{\flushleft \bf{Effect of Weak Labels.}}
We show results for using both pseudo-weak labels as well as human oracle-weak labels. Table \ref{table:gta5_psuedoweak} and \ref{table:synthia_psuedoweak} present the results for different combinations of the modules used in our framework without pixel-level adaptation \cite{Hoffman_ICML_2018}. It is interesting to note that on GTA5$\rightarrow$Cityscapes, even when using pseudo-weak labels, our method obtains a $4.2\%$ boost in performance ($41.4\rightarrow45.6$), as well as a $3-4\%$ boost for SYNTHIA$\rightarrow$Cityscapes. In addition, as expected, using oracle-weak labels performs better than pseudo-weak labels by $6.5\%$ on GTA5$\rightarrow$Cityscapes and $6.5-7.3\%$ on SYNTHIA$\rightarrow$Cityscapes. It it also interesting to note that using the category-wise alignment consistently improves the performance for all the cases, i.e., different types of weak labels and for different datasets.

\begin{figure*}[!t]
    \centering
	\begin{subfigure}{0.49\textwidth}
		\includegraphics[height=5cm]{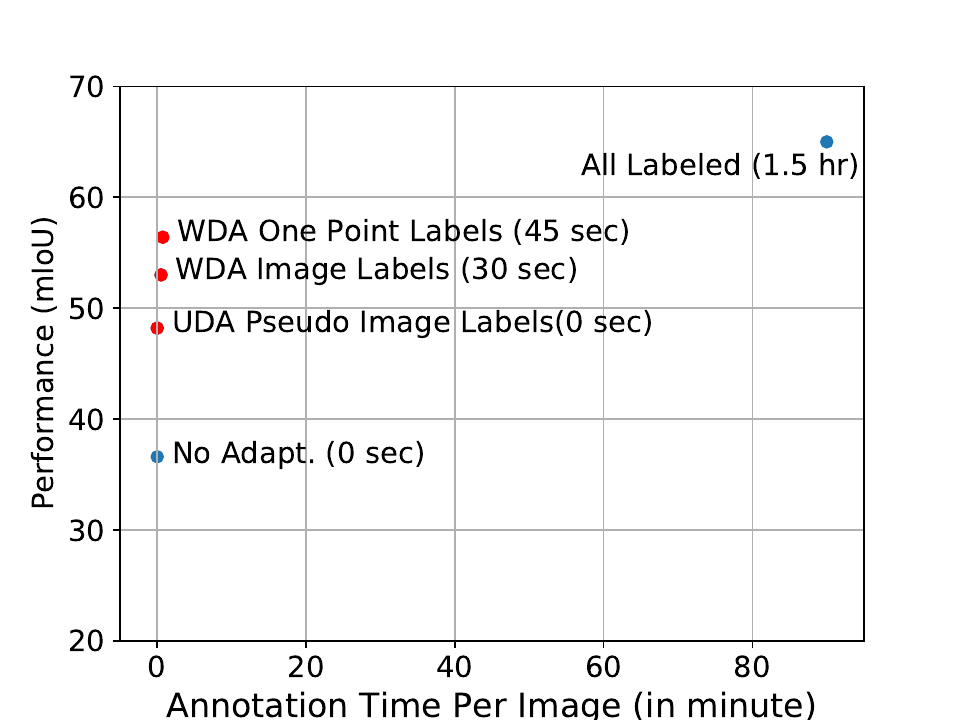}
		\caption{}
		\label{fig:threshold}
	\end{subfigure}
	\begin{subfigure}{0.49\textwidth}
		\includegraphics[height=5cm]{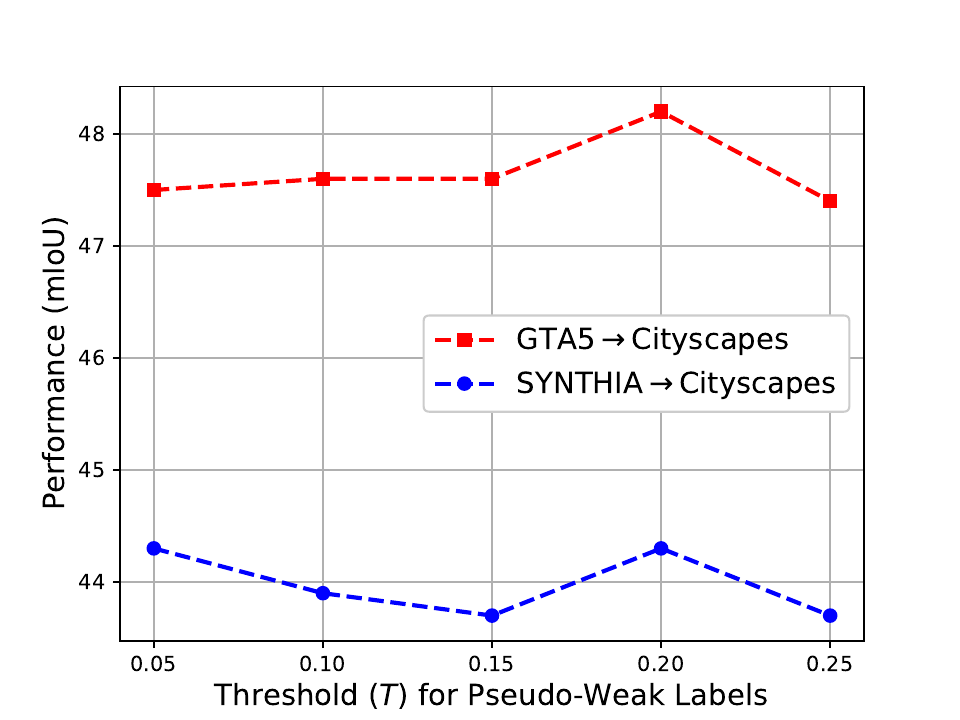}
		\caption{}
		\label{fig:perf_comp}
	\end{subfigure}
	\caption{
     (a) Performance comparison on GTA5$\rightarrow$Cityscapes with different levels of supervision on target images: no target labels (``No Adapt.'' and ``UDA''), weak image labels (30 seconds), one point labels (45 seconds), and fully-supervised setting with all pixels labeled (``All Labeled'') that takes 1.5 hours per image according to \cite{cityscapes}. (b) Performance of our method on GTA5$\rightarrow$Cityscapes with variations in the threshold, i.e., $T$ in \eqref{eq:pseudo_weak}, for obtaining the pseudo-weak labels.}
\end{figure*}

\begin{figure*}[t]
    \centering
    \includegraphics[width=0.95\textwidth]{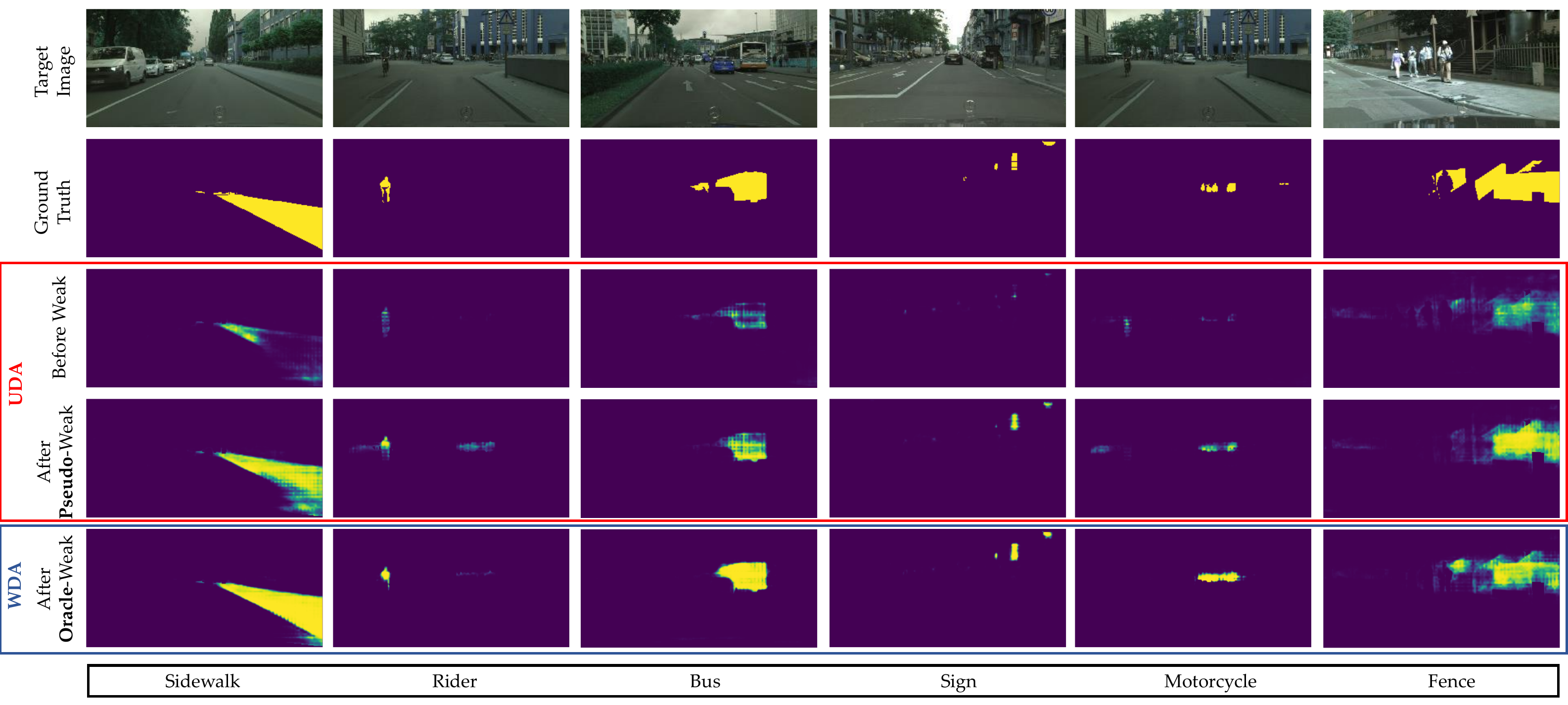}
    \caption{Visualizations of category-wise segmentation prediction probability before and after using the pseudo-weak labels on GTA5$\rightarrow$Cityscapes. Before adaptation, the network only highlights the areas partially with low probability, while using the pseudo-weak labels helps the adapted model obtain much better segments, and is closer to the model using oracle-weak labels.}
    \label{fig:vis}
\end{figure*}

{\flushleft \bf{Effect of Pseudo-Weak Label Threshold.}}
We use a threshold $T$ in \eqref{eq:pseudo_weak} to convert the image-level prediction probability to a  multi-hot vector denoting the pseudo-weak labels that indicates absence/presence of the categories. Note that the threshold is on a probability between $0$ and $1$. We then study the effect of $T$ by varying it and plot the performance in Fig. \ref{fig:threshold} on GTA5$\rightarrow$Cityscapes.
The figure shows that our model generally works well with $T$ in a range of 0.05 to 0.25. However, when we make $T$ larger than $0.3$, the performance starts to drop significantly, as in this case, the recall of the pseudo-weak labels would be very low compared with the oracle-weak labels (i.e., ground truths), which makes the segmentation network fail to predict most categories.

{\flushleft \bf{Output Space Visualization.}}
We present some visualizations of the segmentation prediction probability for each category in Fig. \ref{fig:vis}. Before using any weak labels (third row), the probabilities may be low, even though there is a category present in that image. However, based on these initial predictions, our model can estimate the categories and then enforce their presence/absence explicitly in the proposed classification loss and alignment loss.
The fourth row in Fig. \ref{fig:vis} shows that such pseudo-weak labels help the network discover object/stuff regions towards better segmentation. For example, the fourth and fifth column shows that, although the original prediction probabilities are quite low, results using pseudo-weak labels are estimated correctly. Moreover, the last row shows that the predictions can be further improved when we have oracle-weak labels.

%% file: tables/table_gta5_cityscapes.tex
\begin{table*} [t]
	\caption{Results of adapting GTA5 to Cityscapes.
	The top group is for UDA, while the bottom group presents our method's performance using the oracle-weak labels for WDA that use either image-level or point supervision.
	}
	\label{table:gta5_all}
	\scriptsize
	\centering
	\renewcommand{\arraystretch}{1.7}
	\resizebox{\textwidth}{!}{
	\begin{tabular}{lcccccccccccccccccccc}
		\toprule
		
		& \multicolumn{20}{c}{GTA5 $\rightarrow$ Cityscapes} \\
		\midrule
		
		Method & \rotatebox{90}{road} & \rotatebox{90}{sidewalk} & \rotatebox{90}{building} & \rotatebox{90}{wall} & \rotatebox{90}{fence} & \rotatebox{90}{pole} & \rotatebox{90}{light} & \rotatebox{90}{sign} & \rotatebox{90}{veg} & \rotatebox{90}{terrain} & \rotatebox{90}{sky} & \rotatebox{90}{person} & \rotatebox{90}{rider} & \rotatebox{90}{car} & \rotatebox{90}{truck} & \rotatebox{90}{bus} & \rotatebox{90}{train} & \rotatebox{90}{mbike} & \rotatebox{90}{bike} & mIoU\\
		
		\midrule
		No Adapt. & 75.8 & 16.8 & 77.2 & 12.5 & 21.0 & 25.5 & 30.1 & 20.1 & 81.3 & 24.6 & 70.3 & 53.8 & 26.4 & 49.9 & 17.2 & 25.9 & 6.5 & 25.3 & 36.0 & 36.6 \\
		
		Road~\cite{Chen_CVPR_2018} & 76.3 & 36.1 & 69.6 & 28.6 & 22.4 & 28.6 & 29.3 & 14.8 & 82.3 & 35.3 & 72.9 & 54.4 & 17.8 & 78.9 & 27.7 & 30.3 & 4.0 & 24.9 & 12.6 & 39.4 \\
		
		AdaptOutput~\cite{tsai2018learning} & 86.5 & 25.9 & 79.8 & 22.1 & 20.0 & 23.6 & 33.1 & 21.8 & 81.8 & 25.9 & 75.9 & 57.3 & 26.2 & 76.3 & 29.8 & 32.1 & 7.2 & 29.5 & 32.5 & 41.4 \\
		
		AdvEnt~\cite{Vu_CVPR_2019} & 89.4 & 33.1 & 81.0 & 26.6 & 26.8 & 27.2 & 33.5 & 24.7 & 83.9 & 36.7 & 78.8 & 58.7 & 30.5 & 84.8 & \textbf{38.5} & 44.5 & 1.7 & 31.6 & 32.4 & 45.5 \\
		
		CLAN~\cite{Luo_CVPR_2019} & 87.0 & 27.1 & 79.6 & 27.3 & 23.3 & 28.3 & 35.5 & 24.2 & 83.6 & 27.4 & 74.2 & 58.6 & 28.0 & 76.2 & 33.1 & 36.7 & 6.7 & \textbf{31.9} & 31.4 & 43.2 \\
		
		SWD~\cite{Lee_CVPR_2019} &  \textbf{92.0} & 46.4 & 82.4 & 24.8 & 24.0 & \textbf{35.1} & 33.4 & 34.2 & 83.6 & 30.4 & 80.9 & 56.9 & 21.9 & 82.0 & 24.4 & 28.7 & 6.1 & 25.0 & 33.6 & 44.5 \\
		
		SSF-DAN~\cite{Du_ICCV19} &  90.3 & 38.9 & 81.7 & 24.8 & 22.9 & 30.5 & 37.0 & 21.2 & \textbf{84.8} & 38.8 & 76.9 & 58.8 & 30.7 & \textbf{85.7} & 30.6 & 38.1 & 5.9 & 28.3 & 36.9 & 45.4 \\
		
		DISE~\cite{Chang_CVPR_2019} &  91.5 & 47.5 & 82.5 & 31.3 & 25.6 & 33.0 & 33.7 & 25.8 & 82.7 & 28.8 & 82.7 & \textbf{62.4} & \textbf{30.8} & 85.2 & 27.7 & 34.5 & 6.4 & 25.2 & 24.4 & 45.4 \\
		
		BDL~\cite{Li_CVPR_2019} &  91.4 & 47.9 & \textbf{84.2} & \textbf{32.4} & 26.0 & 31.8 & 37.3 & 33.0 & 83.3 & \textbf{39.2} & 79.2 & 57.7 & 25.6 & 81.3 & 36.3 & 39.7 & 2.6 & 31.3 & 33.5 & 47.2 \\
		
		AdaptPatch~\cite{Tsai_DA4Seg_ICCV19} & 92.3 & \textbf{51.9} & 82.1 & 29.2 & 25.1 & 24.5 & 33.8 & 33.0 & 82.4 & 32.8 & 82.2 & 58.6 & 27.2 & 84.3 & 33.4 & \textbf{46.3} & 2.2 & 29.5 & 32.3 & 46.5 \\
		
		Ours (UDA) & 91.6 & 47.4 & 84.0 & 30.4 & \textbf{28.3} & 31.4 & \textbf{37.4} & \textbf{35.4} & 83.9 & 38.3 & \textbf{83.9} & 61.2 & 28.2 & 83.7 & 28.8 & 41.3 & \textbf{8.8} & 24.7 & \textbf{46.4} & \textbf{48.2} \\
		\midrule
		
		Ours (WDA: Image) & 89.5 & 54.1 & 83.2 & 31.7 & 34.2 & 37.1 & 43.2 & 39.1 & 85.1 & 39.6 & 85.9 & 61.3 & 34.1 & 82.3 & 42.3 & 51.9 & 34.4 & 33.1 & 45.4 & 53.0 \\
		
		Ours (WDA: Point) & 94.0 & 62.7 & 86.3 & 36.5 & 32.8 & 38.4 & 44.9 & 51.0 & 86.1 & 43.4 & 87.7 & 66.4 & 36.5 & 87.9 & 44.1 & 58.8 & 23.2 & 35.6 & 55.9 & 56.4 \\
		
		\bottomrule
	\end{tabular}
	}
\end{table*}

%% file: tables/table_synthia_cityscapes.tex
\begin{table*} [t]
	\caption{
		Results of adapting SYNTHIA to Cityscapes.
		The top group is for UDA, while the bottom group presents the WDA setting using oracle-weak labels. mIoU and mIoU$^\ast$ are averaged over 16 and 13 categories.
	}
	\label{table:synthia_all}
	\scriptsize
	\centering
	\renewcommand{\arraystretch}{1.75}
	\resizebox{\textwidth}{!}{
	\begin{tabular}{lcccccccccccccccccc}
		\toprule
		
		& \multicolumn{18}{c}{SYNTHIA $\rightarrow$ Cityscapes} \\
		\midrule
		
		Method & \rotatebox{90}{road} & \rotatebox{90}{sidewalk} & \rotatebox{90}{building} & \rotatebox{90}{wall} & \rotatebox{90}{fence} & \rotatebox{90}{pole} & \rotatebox{90}{light} & \rotatebox{90}{sign} & \rotatebox{90}{veg} & \rotatebox{90}{sky} & \rotatebox{90}{person} & \rotatebox{90}{rider} & \rotatebox{90}{car} & \rotatebox{90}{bus} & \rotatebox{90}{mbike} & \rotatebox{90}{bike} & mIoU & mIoU$^\ast$ \\

		\midrule
		No Adapt. & 55.6 & 23.8 & 74.6 & 9.2 & 0.2 & 24.4 & 6.1 & 12.1 & 74.8 & 79.0 & 55.3 & 19.1 & 39.6 & 23.3 & 13.7 & 25.0 & 33.5 & 38.6 \\
		
		AdaptOutput~\cite{tsai2018learning} & 79.2 & 37.2 & 78.8 & 10.5 & 0.3 & 25.1 & 9.9 & 10.5 & 78.2 & 80.5 & 53.5 & 19.6 & 67.0 & 29.5 & 21.6 & 31.3 & 39.5 & 45.9 \\
		
		AdvEnt~\cite{Vu_CVPR_2019} & 85.6 & 42.2 & 79.7 & 8.7 & 0.4 & 25.9 & 5.4 & 8.1 & 80.4 & 84.1 & 57.9 & 23.8 & 73.3 & 36.4 & 14.2 & 33.0 & 41.2 & 48.0 \\
		
		CLAN~\cite{Luo_CVPR_2019} &  81.3 & 37.0 & 80.1 & - & - & - & 16.1 & 13.7 & 78.2 & 81.5 & 53.4 & 21.2 & 73.0 & 32.9 & 22.6 & 30.7 &  - & 47.8 \\
		
		SWD~\cite{Lee_CVPR_2019} & 82.4 & 33.2 & \textbf{82.5} & - & - & - & \textbf{22.6} & \textbf{19.7} & \textbf{83.7} & 78.8 & 44.0 &
		 17.9 & 75.4 & 30.2 & 14.4 & 39.9 & - & 48.1 \\
		
		DADA~\cite{Vu_ICCV19} &  89.2 & 44.8 & 81.4 & 6.8 & 0.3 & 26.2 & 8.6 & 11.1 & 81.8 & 84.0 & 54.7 & 19.3 & 79.7 & \textbf{40.7} & 14.0 & 38.8 & 42.6 & 49.8 \\
		
		SSF-DAN~\cite{Du_ICCV19} &  84.6 & 41.7 & 80.8 & - & - & - & 11.5 & 14.7 & 80.8 & \textbf{85.3} & 57.5 & 21.6 & 82.0 & 36.0 & 19.3 & 34.5 & - & 50.0 \\
		
		DISE~\cite{Chang_CVPR_2019} &  91.7 & 53.5 & 77.1 & 2.5 & 0.2 & \textbf{27.1} & 6.2 & 7.6 & 78.4 & 81.2 & 55.8 & 19.2 & \textbf{82.3} & 30.3 & 17.1 & 34.3 & 41.5 & 48.8 \\
		
		
		AdaptPatch~\cite{Tsai_DA4Seg_ICCV19} & 82.4 & 38.0 & 78.6 & 8.7 & \textbf{0.6} & 26.0 & 3.9 & 11.1 & 75.5 & 84.6 & 53.5 & 21.6 & 71.4 & 32.6 & 19.3 & 31.7 & 40.0 & 46.5 \\
		
		Ours (UDA) & \textbf{92.0} & \textbf{53.5} & 80.9 & \textbf{11.4} & 0.4 & 21.8 & 3.8 & 6.0 & 81.6 & 84.4 & \textbf{60.8} & \textbf{24.4} & 80.5 & 39.0 & \textbf{26.0} & \textbf{41.7} & \textbf{44.3} & \textbf{51.9} \\
		\midrule
		Ours (WDA: Image) & 92.3 & 51.9 & 81.9 & 21.1 & 1.1 & 26.6 & 22.0 & 24.8 & 81.7 & 87.0 & 63.1 & 33.3 & 83.6 & 50.7 & 33.5 & 54.7 & 50.6 & 58.5 \\
		
		Ours (WDA: Point) & 94.9 & 63.2 & 85.0 & 27.3 & 24.2 & 34.9 & 37.3 & 50.8 & 84.4 & 88.2 & 60.6 & 36.3 & 86.4 & 43.2 & 36.5 & 61.3 & 57.2 & 63.7 \\
		
		\bottomrule
	\end{tabular}
	}
\end{table*}

%% file: tables/table_ablation.tex
\begin{table}[!t]
		\begin{minipage}[t]{.49\linewidth}
			\scriptsize
			\caption{Ablation of the proposed loss functions for GTA5$\rightarrow$Cityscapes.
    	    }
			\label{table:gta5_psuedoweak}
			\centering
			\renewcommand{\arraystretch}{1.3}
	        \setlength{\tabcolsep}{2pt}
			\begin{tabular}{llcccc}
		    \toprule
		    \multicolumn{5}{c}{GTA5 $\rightarrow$ Cityscapes} \\
		    \midrule
		    & Supervision & $\mathcal{L}_c$ & $\mathcal{L}_{adv}^C$ & mIoU\\
		
		    \midrule
		    
		    \multirow{4}{*}{\rotatebox{90}{UDA}} & No Adapt. & {} & {} & 36.6 \\
		    {} & Baseline \cite{tsai2018learning} & {} & {} & 41.4 \\
		    {} & \multirow{2}{*}{Pseudo-Weak} & {\checkmark} & {} & 44.2 \\
		    {} & {} & {\checkmark} & {\checkmark} & \textbf{45.6} \\
		    \midrule
		    \multirow{2}{*}{\rotatebox{90}{WDA}} & \multirow{2}{*}{Oracle-Weak} & {\checkmark} & {} & 50.8 \\
		    {} & {} & {\checkmark} & {\checkmark} & \textbf{52.1} \\

		    \bottomrule
	    \end{tabular}
		\end{minipage}
		\hfill
		\begin{minipage}[t]{.49\linewidth}
			\scriptsize
			\caption{Ablation of the proposed loss functions for SYNTHIA$\rightarrow$Cityscapes.
        	}
			\label{table:synthia_psuedoweak}
			\centering
			\renewcommand{\arraystretch}{1.3}
    	\setlength{\tabcolsep}{2pt}
    	\begin{tabular}{llccccc}
    		\toprule
    		
    		\multicolumn{6}{c}{SYNTHIA $\rightarrow$ Cityscapes} \\
    		\midrule
    		
    		& Supervision & $\mathcal{L}_c$ & $\mathcal{L}_{adv}^C$ & mIoU & mIoU*\\
    		
    		\midrule
    		
    		\multirow{4}{*}{\rotatebox{90}{UDA}} & No Adapt. & {} & {} & 33.5 & 38.6 \\
    		{} & Baseline \cite{tsai2018learning} & {} & {} & 39.5 & 45.9 \\
    		{} & \multirow{2}{*}{Pseudo-Weak} & {\checkmark} & {} & 41.7 & 49.0 \\
    		{} & {} & {\checkmark} & {\checkmark} & \textbf{42.7} & \textbf{49.9} \\
    		\midrule
    		\multirow{2}{*}{\rotatebox{90}{WDA}} & \multirow{2}{*}{Oracle-Weak} & {\checkmark} & {} & 47.8 & 56.0 \\
    		{} & {} & {\checkmark} & {\checkmark} & \textbf{49.2} & \textbf{57.2} \\

    		\bottomrule
	    \end{tabular}
		\end{minipage}
	\end{table}

%% file: supp.tex
\pagebreak
\begin{center}
\textbf{\Large Supplemental Materials}
\end{center}

\section{Overview}
In this supplementary material, we provide more results of ablation study and using oracle-weak labels, including image-level and point supervisions.
In addition, we present parameter analysis on GTA5 $\rightarrow$ Cityscapes, when using 1) pseudo-weak labels, and 2) image-level oracle-weak labels.
Moreover, we show more visual results of 1) category-wise probability, and 2) final semantic segmentation comparisons. Finally, we also empirically analyze the performance of our model on a different architecture and a different dataset.

\section{Ablation Study}
We provide an ablation study extended from Table 1 and 2 of the main paper.
In Table \ref{table:gta5_psuedoweak} and \ref{table:synthia_psuedoweak}, we add the factor with or without using the pixel-level adaptation (denoted as PA).
From the results, we show that our proposed weak label loss ($\mathcal{L}_c$) and category-wise alignment loss ($\mathcal{L}_{adv}^C$) are both complementary to the PA module, under both the case of UDA and WDA settings.

\begin{table}[h]
		\begin{minipage}[t]{.49\linewidth}
			\scriptsize
			\caption{Ablation of the proposed loss functions for GTA5$\rightarrow$Cityscapes.
    	    }
			\label{table:gta5_psuedoweak}
			\centering
			\renewcommand{\arraystretch}{1.3}
	        \setlength{\tabcolsep}{2pt}
		\begin{tabular}{llccccc}
		\toprule
		
		\multirow{9}{*}{\rotatebox{90}{UDA}} & Supervision & $\mathcal{L}_c$ & $\mathcal{L}_{adv}^C$ & PA & mIoU\\
		
		\midrule
		
		{} & No Adapt. & {} & {} & {} & 36.6 \\
		{} & Baseline \cite{tsai2018learning} & {} & {} & {} & 41.4 \\
		{} & \multirow{4}{*}{Pseudo-Weak} & {\checkmark} & {} & {} & 44.2 \\
		{} & {} & {\checkmark} & {\checkmark} & {} & 45.6 \\
		{} & {} & {\checkmark} & {} & {\checkmark} & 46.7 \\
		{} & {} & {\checkmark} & {\checkmark} & {\checkmark} & \textbf{48.2} \\
		\midrule
		\multirow{4}{*}{\rotatebox{90}{WDA}} & \multirow{4}{*}{Oracle-Weak} & {\checkmark} & {} & {} & 50.8 \\
		{} & {} & {\checkmark} & {\checkmark} & {} & 52.1 \\
		{} & {} & {\checkmark} & {} & {\checkmark} & 52.0 \\
		{} & {} & {\checkmark} & {\checkmark} & {\checkmark} & \textbf{53.0} \\

		\bottomrule
	    \end{tabular}
		\end{minipage}
		\hfill
		\begin{minipage}[t]{.49\linewidth}
			\scriptsize
			\caption{Ablation of the proposed loss functions for SYNTHIA$\rightarrow$Cityscapes.
        	}
			\label{table:synthia_psuedoweak}
			\centering
			\renewcommand{\arraystretch}{1.3}
    	\setlength{\tabcolsep}{2pt}
	\begin{tabular}{llcccccc}
		\toprule

		\multirow{9}{*}{\rotatebox{90}{UDA}} & Supervision & $\mathcal{L}_c$ & $\mathcal{L}_{adv}^C$ & PA & mIoU & mIoU*\\
		
		\midrule
		
		{} & No Adapt. & {} & {} & {} & 33.5 & 38.6 \\
		{} & Baseline \cite{tsai2018learning} & {} & {} & {} & 39.5 & 45.9 \\
		{} & \multirow{4}{*}{Pseudo-Weak} & {\checkmark} & {} & {} & 41.7 & 49.0 \\
		{} & {} & {\checkmark} & {\checkmark} & {} & 42.7 & 49.9 \\
		{} & {} & {\checkmark} & {} & {\checkmark} & 43.0 & 50.6 \\
		{} & {} & {\checkmark} & {\checkmark} & {\checkmark} & \textbf{44.3} & \textbf{51.9} \\
		\midrule
		\multirow{4}{*}{\rotatebox{90}{WDA}} & \multirow{4}{*}{Oracle-Weak} & {\checkmark} & {} & {} & 47.8 & 56.0 \\
		{} & {} & {\checkmark} & {\checkmark} & {} & 49.2 & 57.2 \\
		{} & {} & {\checkmark} & {} & {\checkmark} & 49.8 & 57.8 \\
		{} & {} & {\checkmark} & {\checkmark} & {\checkmark} & \textbf{50.6} & \textbf{58.5} \\

		\bottomrule
	\end{tabular}
		\end{minipage}
	\end{table}

\clearpage

\section{Performance for Oracle-weak Labels}
\begin{figure*}[t!]
    \centering
    \includegraphics[width=0.7\textwidth]{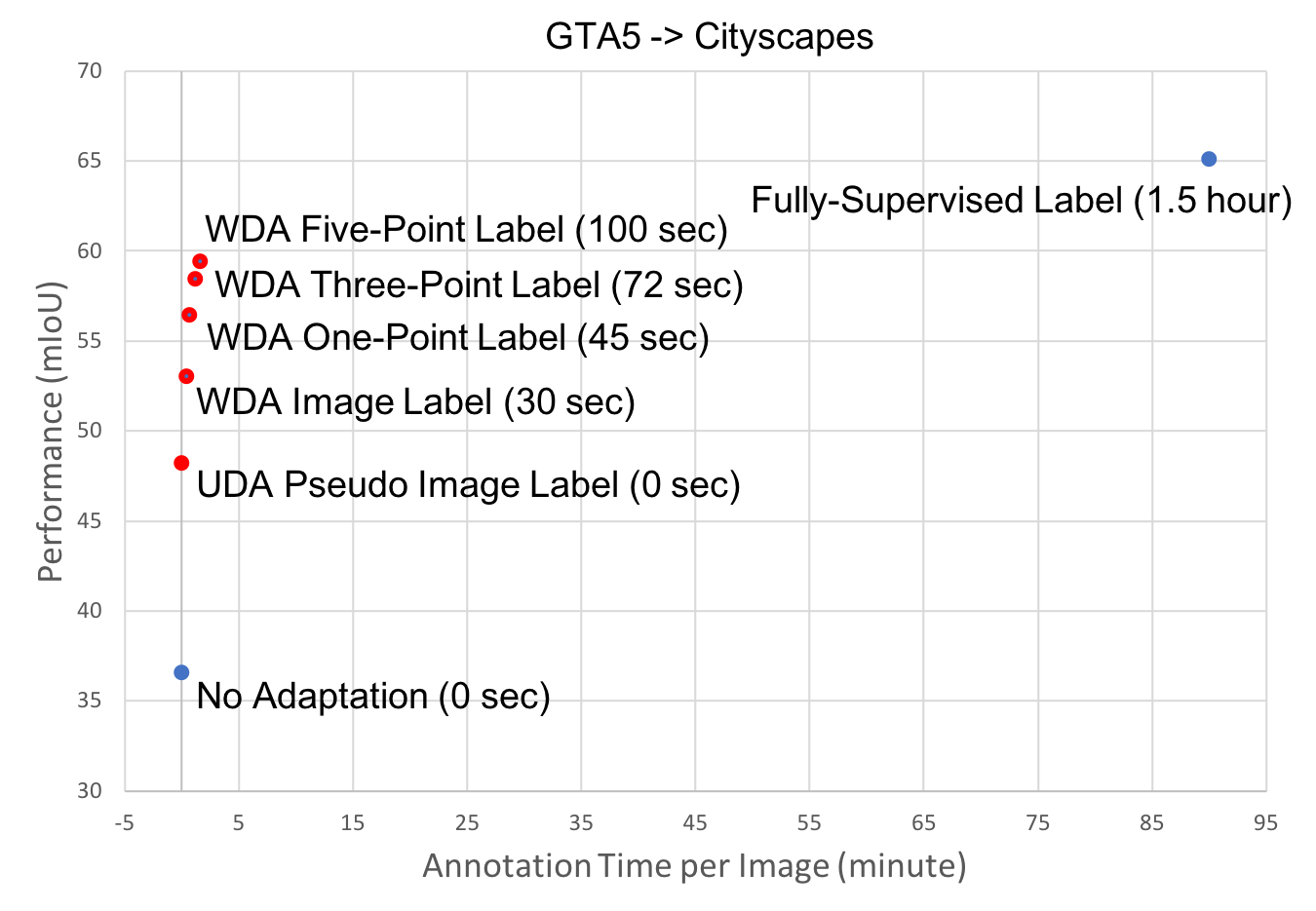}
    \caption{Performance comparison on GTA5 $\rightarrow$ Cityscapes with different levels of supervision on the target domain.}
    \label{fig:time}
\end{figure*}

To further investigate the effectiveness of applying our framework for using oracle-weak labels, we extend the setting of point supervision from annotating 1 point to more points.
In Fig. \ref{fig:time}, we compare the performance v.s. annotation cost on GTA5 $\rightarrow$ Cityscapes, and show that with a small amount of increase in annotation time, the performance can be improved using more points.
For instance, when using 5-point supervision (100 seconds per image), the mIoU reaches 59.4\%, which is close to the fully-supervised setting with 65.1\% mIoU but requiring significantly longer annotation process (1.5 hours per image).
This also demonstrates the usefulness of the proposed novel WDA setting and our framework that can take different types of oracle-weak labels to improve the performance. Note that in experiments, weak labels are extracted from ground truths provided in the dataset. We estimate the annotation time by averaging the time required for a human annotator to label a portion of the dataset.

\section{Parameter Analysis for Pseudo-Weak Labels}
\begin{figure*}[!t]
    \centering
		\begin{subfigure}{0.49\textwidth}
		    \centering
			\includegraphics[width=0.95\textwidth]{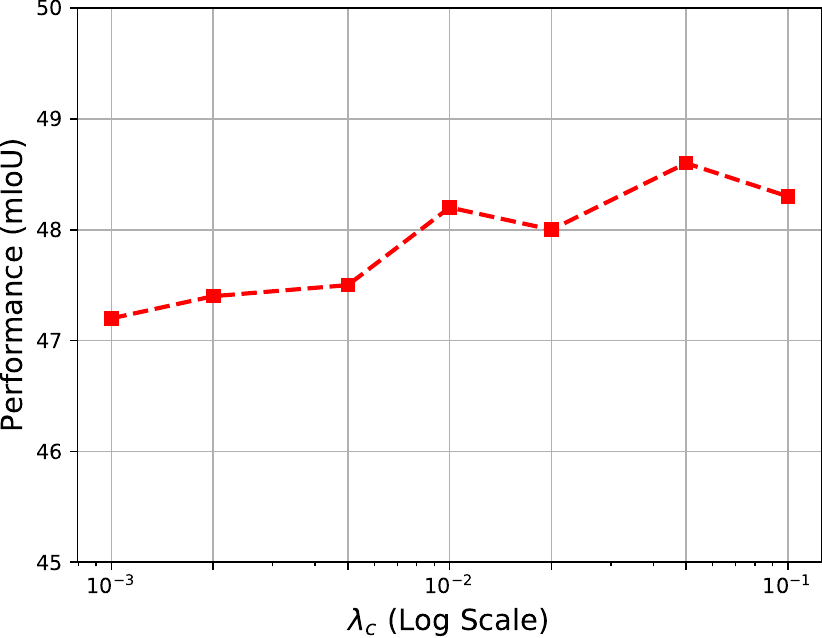}
			\caption{\scriptsize{Weight on Classification Loss}}			
		\end{subfigure}
		\begin{subfigure}{0.49\textwidth}
		    \centering
			\includegraphics[width=0.95\textwidth]{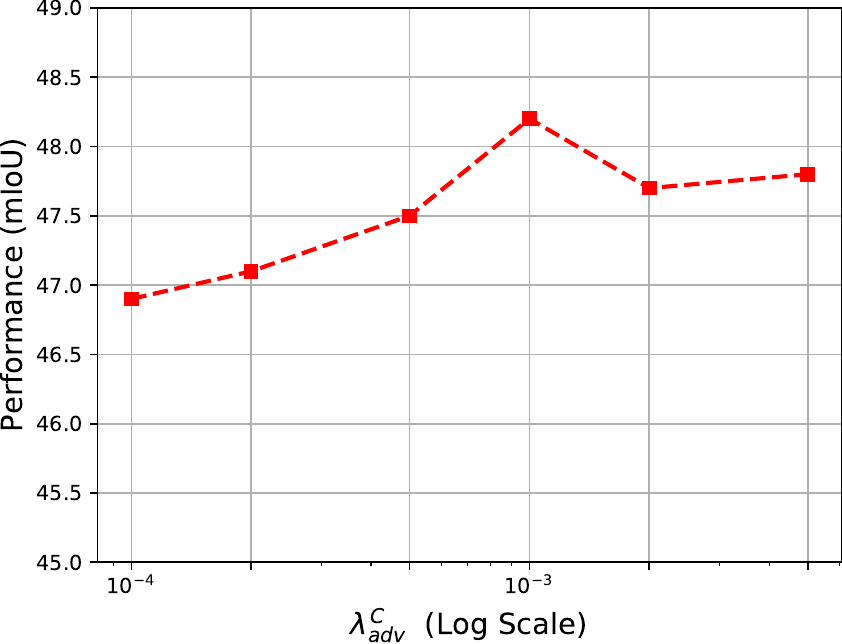}
			\caption{\scriptsize{Weight on Category-wise Alignment Loss}}			
		\end{subfigure}
		\caption{Plots presenting the hyper-parameter analysis of the parameters $\lambda_c$ on the classification loss using pseudo-weak labels and $\lambda_{adv}^C$ on the category-wise alignment loss. }
		\label{fig:pweak_param}
\end{figure*}

Fig. \ref{fig:pweak_param} presents two plots for the parameter analysis when using pseudo-weak labels.
In Fig. \ref{fig:pweak_param}(a), we fix $\lambda_{adv}^C = 0.001$ and show that our model achieves the mIoU larger than $47.5\%$ under a range of $\lambda_c = [0.005, 0.1]$.
When fixing $\lambda_c = 0.01$, Fig. \ref{fig:pweak_param}(b) shows that the model performs well under a range of $\lambda_{adv}^C = [0.0005, 0.005]$. However, when we increase $\lambda_{adv}^C$ to be larger than 0.01, the adversarial training process may become unstable and decreases the performance to $46.2\%$. In addition, decreasing $\lambda_{adv}^C$ would give less focus on alignment and gradually degrades the performance, which shows the importance of our alignment process.

\section{Parameter Analysis for Oracle-Weak Labels}
\begin{figure*}[!t]
    \centering
		\begin{subfigure}{0.49\textwidth}
		    \centering
			\includegraphics[width=0.95\textwidth]{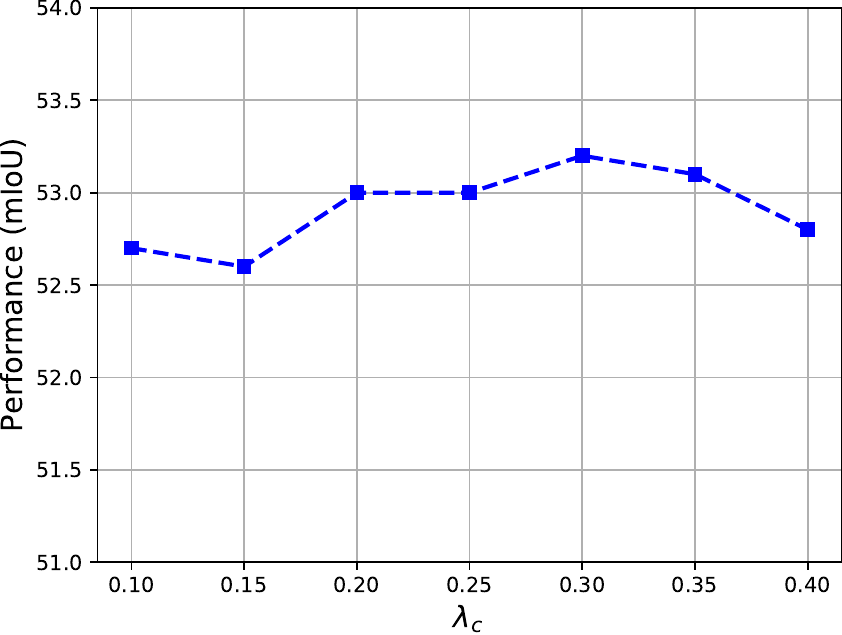}
			\caption{\scriptsize{Weight on Classification Loss}}
			\label{fig:weak_classification}
		\end{subfigure}
		\begin{subfigure}{0.49\textwidth}
		    \centering
			\includegraphics[width=0.95\textwidth]{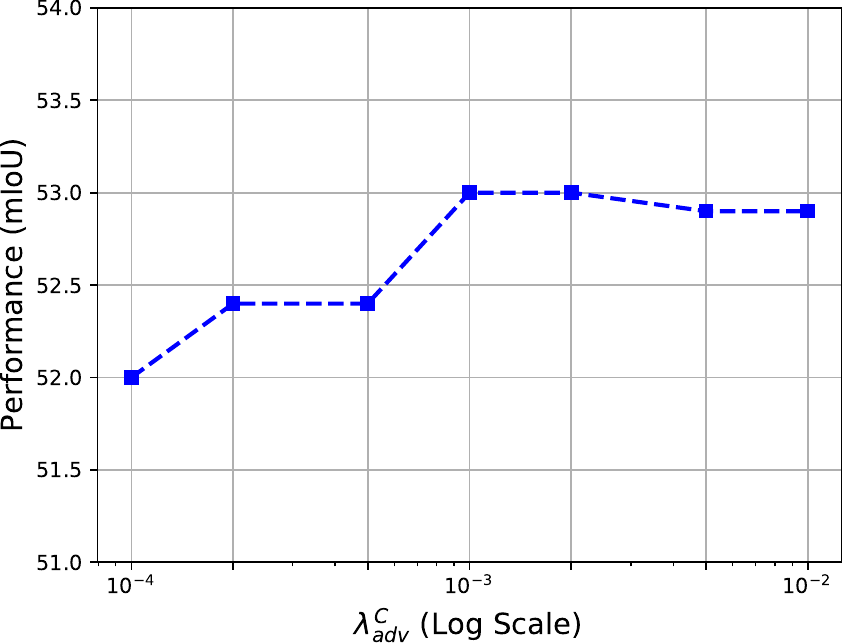}
			\caption{\scriptsize{Weight on Category-wise Alignment Loss}}			
			\label{fig:pweak_alignment}
		\end{subfigure}
		\caption{Plots presenting the hyper-parameter analysis of the parameters $\lambda_c$ on the classification loss using oracle-weak labels and $\lambda_{adv}^C$ on the category-wise alignment loss. }
		\label{fig:weak_param}
\end{figure*}

Fig. \ref{fig:weak_param} presents two plots for the parameter analysis when using image-level oracle-weak labels.
Since such weak labels are accurate, Fig. \ref{fig:weak_param}(a) (fixing $\lambda_{adv} = 0.001$) shows that the performance is quite stable when changing $\lambda_c$.
Moreover, in Fig. \ref{fig:weak_param}(b) fixing $\lambda_c = 0.2$, the performance starts to drop when decreasing $\lambda_{adv}$ as the alignment process becomes weaker.

\section{Category-wise Visualization} 
\begin{figure*}[h]
    \centering
    \includegraphics[width=1.0\textwidth]{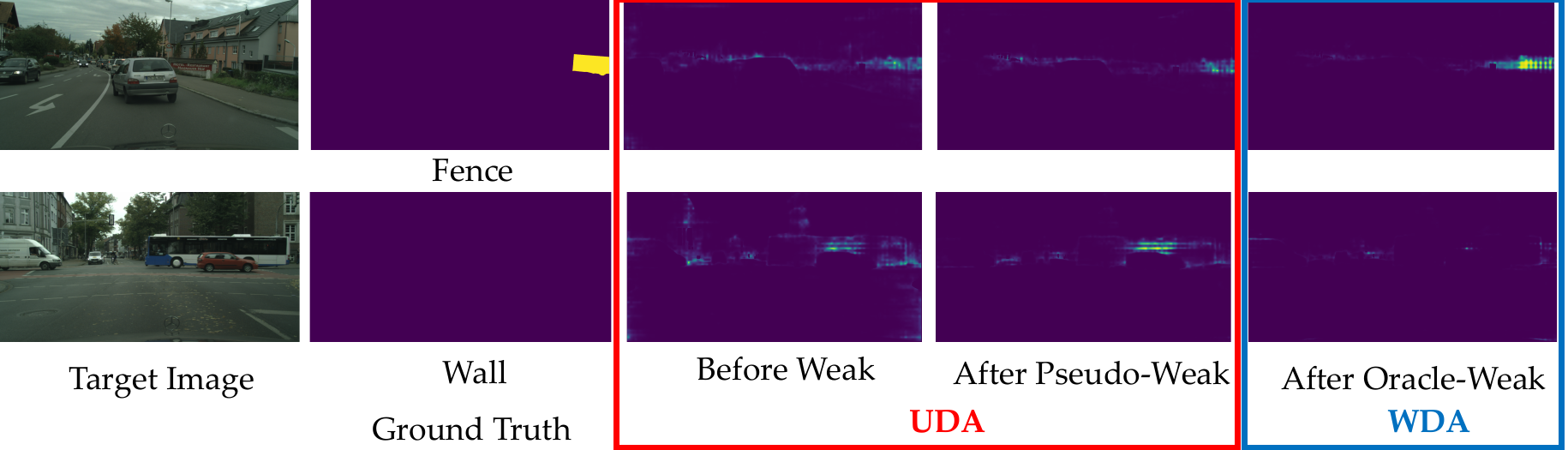}
    \caption{Visualizations for two types of failure cases.}
    \label{fig:failure}
\end{figure*}

\begin{figure*}[h]
    \centering
    \includegraphics[width=1.0\textwidth]{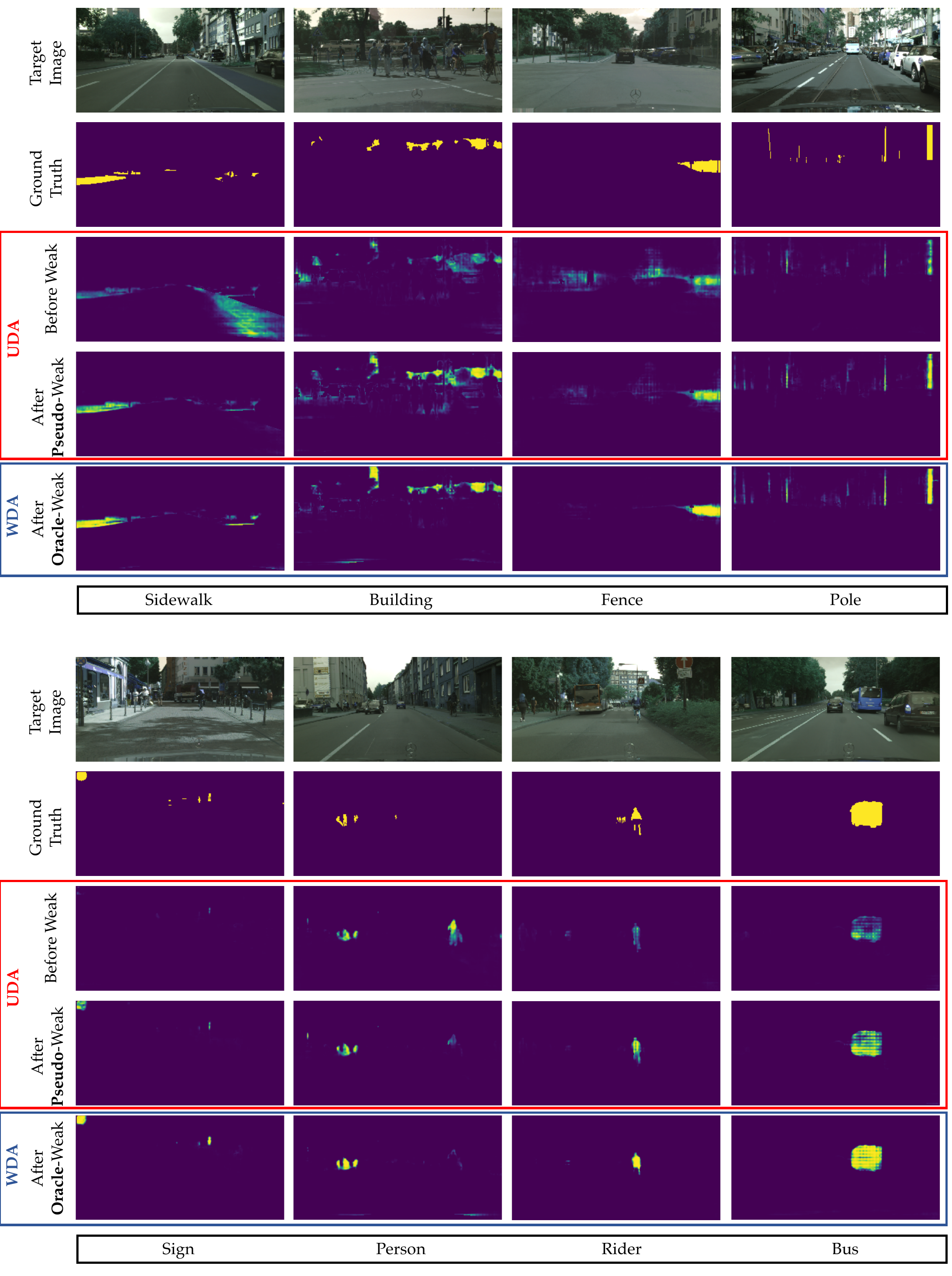}
    \caption{Visualizations of category-wise segmentation prediction probability before and after using the pseudo-weak labels on
GTA5 $\rightarrow$ Cityscapes. Before adaptation, the network only highlights the areas partially with low probability, while using the pseudo-weak labels helps the adapted model obtain much better segments, and is closer to the model using oracle-weak labels.}
    \label{fig:vis_suppl}
\end{figure*}

Fig. \ref{fig:vis_suppl} presents some example visualizations showing the category-wise spatial probabilities before using any weak labels for adaptation, after using pseudo-weak labels and finally after using oracle-weak labels for adaptation.
For example, before using weak labels (e.g., \textit{Sidewalk} and \textit{Fence}), some regions are highlighted incorrectly, and those regions are eliminated after using pseudo-weak labels (UDA) and oracle-weak labels (WDA).
In addition, even with a small portion being highlighted before using weak labels (e.g., \textit{Sign}, \textit{Rider}, \textit{Bus}), the probability maps become more prominent after using weak labels.
Moreover, we present two failure cases of the pseudo-weak labels in Fig. \ref{fig:failure}. In the first row, category \textit{Fence} occurs, but is not detected in the pseudo-weak label. In the second row, category \textit{Wall} does not appear, but is detected in the pseudo-weak label.

\clearpage

\begin{figure*}[h]
    \centering
    \includegraphics[width=1.0\textwidth]{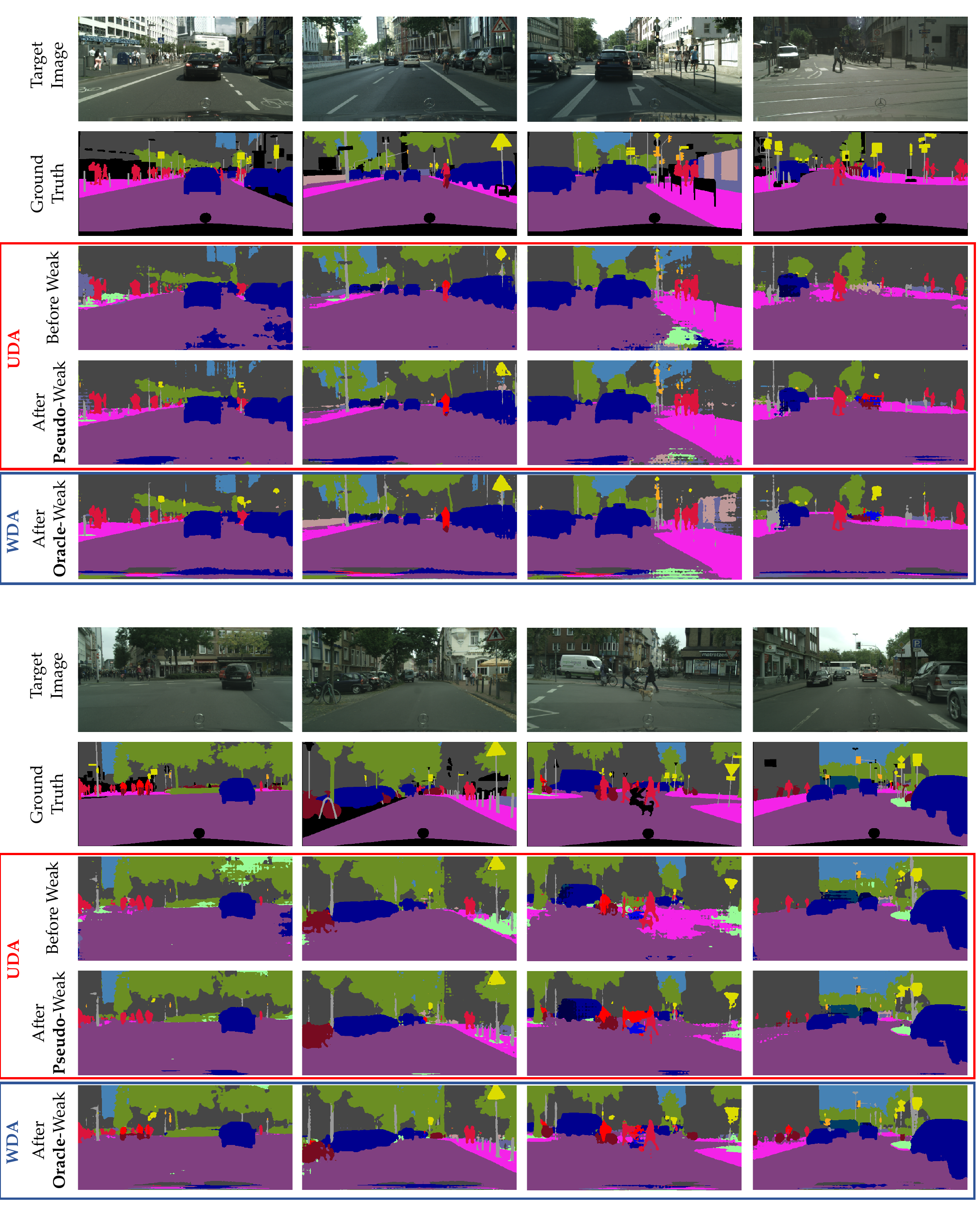}
    \caption{Example results of adapted segmentation for GTA5 $\rightarrow$ Cityscapes with and without using weak labels for adaptation. The visualizations show that using pseudo-weak labels, the segmentation become more structured and some of the categories are better segmented. Using oracle-weak labels further improves the segmentation quality.}
    \label{fig:vis_suppl_color}
\end{figure*}

\section{Semantic Segmentation Visualization}
Fig. \ref{fig:vis_suppl_color} presents the semantic segmentation results before and after using weak labels for adaptation. The UDA method without using any weak labels produces more erroneous results in some portions and may miss some of the categories within a small area, such as sign, pole, etc. However, using the pseudo-weak labels enhances the segmentation and helps our model better identify the categories which originally have a lower confidence. Moreover, using oracle-weak labels is able to further improve the segmentation performance.

\section{More Results on Architecture and Dataset}
Table \ref{table:gta5_vgg} presents segmentation performance using the VGG16 architecture with GTA5 as source and Cityscapes as target. Our method performs better than other UDA methods. We also present results for the WDA case with oracle-weak labels, i.e., image or point labels, which produces higher performance than the UDA methods.

Moreover, we test our method with GTA5 as source and Foggy Cityscapes \cite{sakaridis2018semantic} as target. There is a parameter to choose the level of fog in the images, and we set that to 0.02 in our experiments. The results are presented in Table \ref{table:gta5_foggy}. We can observe consistent improvements as in other datasets.

\begin{table*}[h]
	\caption{Results of adapting GTA5 to Cityscapes with VGG16. The top group is for Unsupervised Domain Adaptation (UDA), while the bottom group presents our method’s performance using the oracle-weak labels for Weakly-supervised Domain Adaptation (WDA) that use either image-level or point supervision.}
	\label{table:gta5_vgg}
	\scriptsize
	\centering
	\renewcommand{\arraystretch}{1.7}
	\resizebox{\textwidth}{!}{
	\begin{tabular}{lcccccccccccccccccccc}
		\toprule
		
		& \multicolumn{20}{c}{GTA5 $\rightarrow$ Cityscapes} \\
		\midrule
		
		Method & \rotatebox{90}{road} & \rotatebox{90}{sidewalk} & \rotatebox{90}{building} & \rotatebox{90}{wall} & \rotatebox{90}{fence} & \rotatebox{90}{pole} & \rotatebox{90}{light} & \rotatebox{90}{sign} & \rotatebox{90}{veg} & \rotatebox{90}{terrain} & \rotatebox{90}{sky} & \rotatebox{90}{person} & \rotatebox{90}{rider} & \rotatebox{90}{car} & \rotatebox{90}{truck} & \rotatebox{90}{bus} & \rotatebox{90}{train} & \rotatebox{90}{mbike} & \rotatebox{90}{bike} & mIoU\\
		
		\midrule

		AdaptOutput~\cite{tsai2018learning} & 87.3 & 29.8 & 78.6 & 21.1 & 18.2 & 22.5 & 21.5 & 11.0 & 79.7 & 29.6 & 71.3 & 46.8 & 6.5 & 80.1 & 23.0 & 26.9 & 0.0 & 10.6 & 0.3 & 35.0 \\
		
		AdvEnt~\cite{Vu_CVPR_2019} & 86.9 & 28.7 & 78.7 & 28.5 & \textbf{25.2} & 17.1 & 20.3 & 10.9 & 80.0 & 26.4 & 70.2 & 47.1 & 8.4 & 81.5 & 26.0 & 17.2 & \textbf{18.9} & 11.7 & 1.6 & 36.1 \\

		CLAN~\cite{Luo_CVPR_2019} & 88.0 & 30.6 & 79.2 & 23.4 & 20.5 & \textbf{26.1} & 23.0 & \textbf{14.8} & 81.6 & \textbf{34.5} & 72.0 & 45.8 & 7.9 & 80.5 & \textbf{26.6} & \textbf{29.9} & 0.0 & 10.7 & 0.0 & 36.6 \\
		
		SSF-DAN~\cite{Du_ICCV19} &  \textbf{88.7} & 32.1 & \textbf{79.5} & 29.9 & 22.0 & 23.8 & 21.7 & 10.7 & 80.8 & 29.8 & \textbf{72.5} & 49.5 & 16.1 & \textbf{82.1} & 23.2 & 18.1 & 3.5 & \textbf{24.4} & 8.1 & 37.7 \\
		
		AdaptPatch~\cite{Tsai_DA4Seg_ICCV19} & 87.3 & \textbf{35.7} & \textbf{79.5} & \textbf{32.0} & 14.5 & 21.5 & 24.8 & 13.7 & 80.4 & 32.0 & 70.5 & 50.5 & 16.9 & 81.0 & 20.8 & 28.1 & 4.1 & 15.5 & 4.1 & 37.5 \\
		
		Ours (UDA) & 87.1 & \textbf{35.7} & 78.6 & 24.9 & 22.7 & 21.8 & \textbf{26.5} & 11.7 & \textbf{82.1} & 32.1 & 70.4 & \textbf{50.6} & \textbf{18.3} & 77.4 & 21.7 & 24.6 & 7.6 & 16.3 & \textbf{19.3} &  \textbf{38.4} \\
		\midrule
		
		Ours (Image) & 88.0 & 46.8 & 81.6 & 22.3 & 35.2 & 27.4 & 29.2 & 27.0 & 82.4 & 35.4 & 80.7 & 57.1 & 29.0 & 83.2 & 38.0 & 56.4 & 23.3 & 29.8 & 5.5 & 46.2 \\
		
		Ours (Point) & 93.6 & 62.7 & 81.4 & 29.6 & 33.7 & 30.7 & 29.7 & 38.2 & 81.5 & 43.0 & 81.7 & 54.3 & 28.8 & 83.8 & 42.9 & 52.5 & 38.4 & 27.1 & 49.8 & 51.8 \\
		
		\bottomrule
	\end{tabular}
	}
\end{table*}

\begin{table*}[h]
	\caption{Results of adapting GTA5 to Foggy Cityscapes with ResNet101. The top group is for Unsupervised Domain Adaptation (UDA), while the bottom group presents our method’s performance using the oracle-weak labels for Weakly-supervised Domain Adaptation (WDA) that use either image-level or point supervision.}
	\label{table:gta5_foggy}
	\scriptsize
	\centering
	\renewcommand{\arraystretch}{1.7}
	\resizebox{\textwidth}{!}{
	\begin{tabular}{lcccccccccccccccccccc}
		\toprule
		
		& \multicolumn{20}{c}{GTA5 $\rightarrow$ Cityscapes} \\
		\midrule
		
		Method & \rotatebox{90}{road} & \rotatebox{90}{sidewalk} & \rotatebox{90}{building} & \rotatebox{90}{wall} & \rotatebox{90}{fence} & \rotatebox{90}{pole} & \rotatebox{90}{light} & \rotatebox{90}{sign} & \rotatebox{90}{veg} & \rotatebox{90}{terrain} & \rotatebox{90}{sky} & \rotatebox{90}{person} & \rotatebox{90}{rider} & \rotatebox{90}{car} & \rotatebox{90}{truck} & \rotatebox{90}{bus} & \rotatebox{90}{train} & \rotatebox{90}{mbike} & \rotatebox{90}{bike} & mIoU\\
		
		\midrule
		No Adapt. & 78.8 & 11.8 & 67.8 & 15.1 & 15.6 & 19.5 & 20.6 & 12.1 & 63.6 & 19.3 & 60.3 & 49.3 & 22.6 & 55.6 & 17.2 & 14.9 & 0.0 & 19.2 & 27.0 & 31.0 \\

		AdaptOutput~\cite{tsai2018learning} & 87.3 & 24.9 & 70.2 & 15.4 & 18.7 & 19.6 & 24.9 & 18.6 & \textbf{69.3} & 28.2 & 64.4 & 49.5 & 24.1 & 74.0 & \textbf{17.6} & 21.2 & 2.1 & \textbf{27.5} & 35.9 & 36.5 \\
		
		Ours (UDA) & \textbf{88.8} & \textbf{27.8} & \textbf{71.0} & \textbf{21.7} & \textbf{21.8} & \textbf{26.4} & \textbf{33.1} & \textbf{26.2} &  68.7 & \textbf{29.4} & \textbf{66.3} & \textbf{55.4} & \textbf{27.2} & \textbf{77.1} & 11.8 & \textbf{24.0} & \textbf{5.7} & 14.7 & \textbf{39.3} & \textbf{38.8} \\
		\midrule
		
		Ours (Image) & 89.0 & 32.8 & 76.5 & 22.0 & 26.5 & 29.8 & 35.3 & 34.8 & 77.4 & 32.8 & 71.7 & 60.1 & 35.0 & 84.7 & 33.6 & 42.0 & 19.0 & 30.8 & 44.1 & 46.2 \\
		
		Ours (Point) & 92.7 & 55.0 & 80.0 & 28.3 & 29.3 & 34.2 & 37.4 & 45.8 & 79.9 & 32.8 & 73.4 & 62.4 & 34.0 & 85.8 & 37.2 & 50.6 & 19.3 & 28.1 & 53.7 & 50.5 \\
		
		\bottomrule
	\end{tabular}
	}
\end{table*}